\documentclass[journal,twoside,web]{ieeecolor}
\usepackage{generic}
\usepackage{cite}
\usepackage{amsmath,amssymb,amsfonts}
\usepackage{algorithmic}
\usepackage{graphicx}
\usepackage{algorithm,algorithmic}
\usepackage{hyperref}
\usepackage{caption} 
\usepackage{subcaption} 
\usepackage{booktabs}
\usepackage{multirow}
\usepackage{array} 
\hypersetup{hidelinks=true}
\usepackage{textcomp}

\def\BibTeX{{\rm B\kern-.05em{\sc i\kern-.025em b}\kern-.08em
    T\kern-.1667em\lower.7ex\hbox{E}\kern-.125emX}}
\begin{document}
\title{DualStabSleepNet: A Dual-Domain Diffusion Stabilization Network for Robust Sleep Staging}
\author{Chongjian Wang, Chen Liu, Junjie Gao, Xiaofang Zhong, \\
Shiyuan Han, and Tong Zhang.,\IEEEmembership{Senior Member, IEEE}
\thanks{Chongjian Wang and Chen Liu are with the School of Mathematics and Systems Science, Shandong University of Science and Technology, Qingdao, 266590, Shandong, China (e-mail: 202311080223@sdust.edu.cn, 202311081110@sdust.edu.cn)}
\thanks{Junjie Gao, Xiaofang Zhong and Shiyuan Han are with the School of Artificial Intelligence, Shandong Women’s University, Jinan, 250352, Shandong, China
(e-mail: junjie.gao@sdwu.edu.cn, 34046@sdwu.edu.cn, ai$\_$hansy@sdwu.edu.cn)}
\thanks{Tong Zhang is with the School of Computer Science and Engineering, South China University of Technology, Guangzhou, 510641, Guangdong, China(e-mail: tony@scut.edu.cn)}
\thanks{Corresponding author: Junjie Gao.}}

\maketitle

\begin{abstract}
In automatic sleep staging, existing deep learning methods often exhibit limited robustness under heterogeneous recording conditions, where non-stationary noise, inter-subject variability, and cross-dataset distribution shifts lead to unstable feature representations and degraded generalization performance. To address these issues, this study proposes DualStabSleepNet (DSSNet), a dual-domain diffusion stabilization network that explicitly enhances robustness at both the data domain and the feature representation domain. Through multi-channel polysomnography (PSG) data input and preprocessing, DSSNet first employs a continuous-scale diffusion-based stabilization module to suppress noise while preserving physiologically meaningful signal structures. The stabilized signals are then transformed into time–frequency representations and encoded by a Vision Transformer backbone. To further alleviate representation drift, DSSNet integrates diffusion-based feature stabilization modules guided by a teacher–student mechanism, reinforcing structural consistency across multiple feature hierarchies. Extensive experiments are conducted on four public PSG benchmarks, including SleepEDF-20, SleepEDF-78, SHHS, and ISRUC-S3. DSSNet achieves state-of-the-art classification accuracies of 89.2\%, 88.0\%, 89.7\%, and 86.7\% on these datasets, respectively, while consistently improving macro-F1 and Cohen's $\kappa$ over the strongest competing methods. Notably, DSSNet yields substantial gains on challenging transitional stages such as N1 (up to 12.5\% improvement on SHHS) and significantly enhances N2 and REM recognition. Moreover, under strict cross-dataset evaluation protocols, DSSNet maintains strong robustness against distribution shifts and achieves performance comparable to or exceeding multiple models trained directly on the target datasets, demonstrating its practical value for reliable real-world sleep staging across heterogeneous cohorts.

\end{abstract}


\section{Introduction}
\label{sec:introduction}

\IEEEPARstart{S}{leep} staging is a fundamental component of sleep medicine and sleep health assessment, serving as a cornerstone for diagnosing sleep disorders, evaluating sleep quality, and supporting long-term personalized healthcare [1]. Polysomnography (PSG), which synchronously records multiple physiological signals including electroencephalography (EEG), electrooculography (EOG), and electromyography (EMG), remains the clinical gold standard for sleep stage annotation [2], [3]. However, manual epoch-wise scoring of PSG recordings is time-consuming, subjective, and difficult to scale in large clinical and home-monitoring settings, thereby motivating the development of reliable automatic sleep staging systems [4].

In recent years, machine learning and deep learning approaches have achieved substantial progress in automatic sleep staging. Nevertheless, robust deployment in real-world environments remains challenging. PSG signals are inherently non-stationary and are frequently corrupted by noise, motion artifacts, and channel-quality fluctuations [5], [6]. Moreover, pronounced inter-subject physiological variability further complicates the learning of stable discriminative patterns [7], [8]. Beyond subject-level differences, substantial distribution shifts arise across datasets due to variations in acquisition devices, channel configurations, sampling rates, and cohort characteristics [8], [9]. As a consequence, models that perform well under controlled within-dataset evaluations often suffer notable performance degradation in cross-subject and, more critically [8], [9], cross-dataset scenarios. Such limitations hinder practical adoption in heterogeneous clinical environments where consistent performance is essential.

From a mechanistic standpoint, the above performance degradation cannot be solely attributed to insufficient classifier capacity. Instead, it reflects coupled sources of instability emerging across multiple domains. At the data (signal) domain, noise contamination and acquisition variability reduce input fidelity and undermine the reliability of downstream feature extraction [10], [11]. At the feature (representation) domain, samples belonging to the same sleep stage may exhibit substantial dispersion and drift in high-dimensional feature space across subjects and datasets, resulting in fragile decision boundaries and weakened generalization [7], [12]. Importantly, data-domain instability primarily affects the quality of physiological observations, whereas feature-domain instability directly governs the robustness of discriminative structures. However, most existing methods address robustness only implicitly [9], [13], often as a by-product of stronger backbones, data augmentation, or domain adaptation, without explicitly modeling stabilization as a first-class objective across both domains.

Motivated by these observations, this work revisits automatic sleep staging from a stabilization-oriented perspective and proposes \emph{DualStabSleepNet} (DSSNet), a dual-domain diffusion stabilization network designed to enhance robustness under heterogeneous recording conditions. DSSNet explicitly targets instability at both the data domain and the feature representation domain within a unified continuous-scale diffusion framework. In the data domain, a diffusion-based stabilization module performs adaptive and structure-preserving enhancement, suppressing non-stationary noise while retaining physiologically meaningful sleep-related patterns. In the feature domain, a diffusion-guided feature stabilization mechanism mitigates representation drift and reinforces structural consistency across hierarchical feature spaces, thereby improving the stability of discriminative boundaries. Notably, DSSNet adopts a coherent modeling paradigm across domains without introducing additional inference complexity.

We evaluate DSSNet on multiple publicly available PSG datasets under strict cross-dataset training--testing protocols. Experimental results demonstrate that DSSNet achieves competitive overall classification performance while significantly improving generalization stability across heterogeneous conditions. In particular, consistent gains are observed for challenging and transitional sleep stages such as N1, highlighting the practical value of explicit dual-domain stabilization for real-world sleep staging applications.

The main contributions of this work are summarized as follows:
\begin{itemize}
\item We propose DSSNet, a dual-domain diffusion stabilization framework that explicitly enhances robustness at both the data domain and the feature domain for automatic sleep staging under heterogeneous recording conditions.
\item We develop a diffusion-guided feature stabilization mechanism that reinforces structural consistency in high-dimensional feature space and mitigates representation drift across subjects and datasets.
\item We conduct systematic cross-dataset evaluations on multiple public PSG benchmarks, demonstrating improved generalization stability and more reliable recognition of difficult transitional sleep stages.
\end{itemize}

\section{Related Work}

{\textbf{Machine Learning-Based Methods.}}
Automatic sleep staging has been extensively investigated over the past decades. Early studies primarily relied on conventional machine learning pipelines, typically following a two-stage paradigm of handcrafted feature extraction and shallow classification. Features were manually designed from EEG, EOG, and EMG signals in the time domain (e.g., statistical descriptors and Hjorth parameters), frequency domain (e.g., power spectral density and band-energy ratios), and time--frequency domain (e.g., wavelet coefficients), sometimes complemented by nonlinear dynamical measures. These engineered representations were then fed into classifiers such as support vector machines (SVMs) or random forests (RFs) [14] ,[15]. Although such approaches provide a certain degree of physiological interpretability, their performance is strongly constrained by the completeness and discriminability of handcrafted features. Consequently, they are highly sensitive to noise contamination, inter-subject variability, and heterogeneous acquisition conditions, limiting their robustness and scalability in large clinical cohorts.

\vspace{2pt}
{\textbf{Deep Learning-Based Methods.}}
With the advent of deep learning, end-to-end representation learning has become the dominant paradigm for automatic sleep staging. Convolutional neural networks (CNNs) and their variants, including DeepSleepNet [16], TinySleepNet [17], AttnSleep [18], DilatedSleepNet [19], and SleepEEGNet [13], have demonstrated strong capability in capturing spectral--temporal patterns through multi-scale convolutional structures and attention mechanisms. Hybrid CNN--RNN frameworks such as SeqSleepNet [20], XSleepNet [12], and MVF-SleepNet [21] further incorporate recurrent units to model inter-epoch temporal dependencies and sleep stage transition dynamics. More recently, Transformer-based architectures have been introduced to better capture long-range contextual relationships across overnight recordings, often combined with CNN front-ends or explicit multi-channel fusion strategies [22], [23]. In parallel, graph neural networks and spatial modeling approaches have also been explored to encode inter-channel relationships in multi-channel EEG recordings [24].

Despite these advances, most deep learning-based methods mainly optimize discriminative objectives and achieve strong performance under within-dataset or controlled evaluation settings. However, growing evidence suggests that their generalization ability remains limited in cross-subject and, more critically, cross-dataset scenarios [25], [26]. To mitigate performance degradation caused by distribution shifts, prior works have explored data augmentation, domain adaptation, adversarial learning, and signal preprocessing techniques such as filtering and denoising [27]. Feature-level regularization and alignment strategies have also been proposed to enhance robustness. Nevertheless, these approaches often assume that distribution discrepancies are relatively static or can be corrected in a one-shot manner, which restricts their effectiveness under continuously varying noise conditions and complex real-world PSG acquisition environments. Moreover, robustness is typically addressed in a single domain---either at the signal level or the feature level---without explicitly modeling the coupled interaction between signal degradation and representation drift, leaving fundamental stability challenges insufficiently resolved.

\vspace{2pt}
{\textbf{Diffusion Models.}}
Diffusion models have recently gained increasing attention in physiological signal analysis due to their strong capability in modeling stochastic noise processes while preserving structural information [27], [28]. Existing studies have investigated diffusion-based approaches for EEG or ECG denoising, enhancement, and signal reconstruction, where noisy observations are progressively mapped back to physiologically consistent signals through learned reverse diffusion processes. While these methods demonstrate promising potential in improving signal quality, their application to automatic sleep staging remains relatively limited. Most diffusion-related works treat diffusion models primarily as standalone signal enhancement modules, with limited integration into downstream discriminative representation learning [27], [28], [29], [30]. Furthermore, existing studies largely focus on waveform-level restoration, whereas the potential of diffusion mechanisms for stabilizing high-dimensional feature representations---and their role in improving classification robustness across heterogeneous datasets---has received little exploration.

\vspace{2pt}
In contrast to the above approaches, this work explicitly treats stability as a first-class modeling objective for automatic sleep staging under heterogeneous recording conditions. By jointly addressing signal-domain degradation and representation-domain drift within a unified diffusion-based stabilization framework, the proposed method bridges the gap between physiological signal enhancement and robust discriminative learning. This dual-domain design enables complementary stabilization in both the data domain and the feature domain, providing a principled and computationally efficient solution for improving generalization consistency in challenging cross-subject and cross-dataset scenarios.

\section{Method}
\begin{figure*}[!t]
    \centering
    \includegraphics[width=1\textwidth]{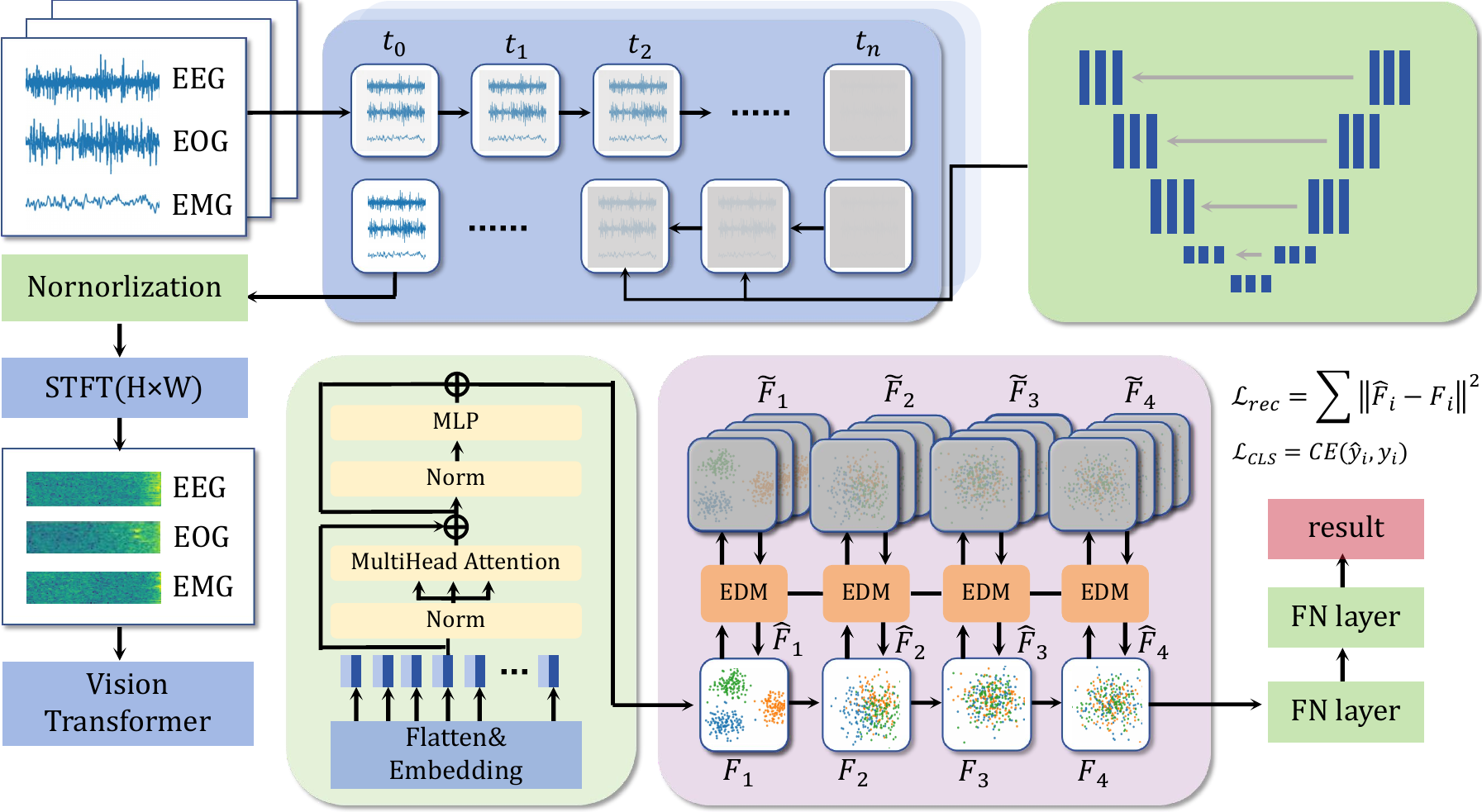}
    \caption{(a) Data domain. Raw multi-channel PSG signals (EEG, EOG, and EMG) are normalized and processed by a continuous-scale diffusion process (blue), where a U-Net–based denoiser (green) progressively suppresses noise while preserving physiological signal structures.
(b) Feature domain. The enhanced 1D signals are transformed into 2D time–frequency representations via STFT, encoded by a Vision Transformer (green), and further stabilized by feature-domain diffusion modules (purple) before feature fusion and sleep stage classification.}
    \label{fig:overall}
\end{figure*}

\subsection{A Unified Continuous-Scale Diffusion Framework}

Polysomnographic (PSG) signals used for sleep staging are non-stationary and frequently corrupted by artifacts whose intensity varies continuously over time. Meanwhile, sleep-related physiological structures often overlap with ocular and muscular artifacts in both time and frequency domains [27], [28], [29], which limits the effectiveness of fixed-parameter filtering and denoising methods based on discrete noise assumptions. To address this issue, we model signal degradation as a continuous-scale perturbation process and learn to recover stable structures under arbitrary noise strengths. As illustrated in Fig.~\ref{fig:overall}, this continuous-scale diffusion formulation serves as the common modeling backbone of DualStabSleepNet (DSSNet), enabling unified stabilization in both the data domain and the feature representation domain.

Let $x_{\mathrm{clean}}$ denote a latent stable target, which may correspond to a physiologically meaningful waveform or a statistically stable representation. The forward perturbation process is defined as
\begin{equation}
x = x_{\mathrm{clean}} + \sigma \varepsilon, \quad 
\varepsilon \sim \mathcal{N}(0, I), \quad 
\sigma \in [\sigma_{\min}, \sigma_{\max}],
\end{equation}
where $\sigma$ is a continuously sampled noise scale controlling the perturbation strength. Unlike discrete diffusion formulations, this definition does not quantize noise levels but models degradation as a smooth continuum.

For reverse modeling, we adopt the data-prediction parameterization, in which the denoiser directly regresses the stable target:
\begin{equation}
D_{\theta}(x, \sigma) \approx x_{\mathrm{clean}} .
\end{equation}
The model is trained by minimizing a scale-weighted mean squared error objective:
\begin{equation}
\mathcal{L}(\theta) =
\mathbb{E}_{x_{\mathrm{clean}}, \sigma, \varepsilon}
\left[
w(\sigma)
\left\|
D_{\theta}(x_{\mathrm{clean}} + \sigma \varepsilon, \sigma)
- x_{\mathrm{clean}}
\right\|_2^2
\right],
\end{equation}
whose optimal solution satisfies
\begin{equation}
D^{*}(x, \sigma) = \mathbb{E}[x_{\mathrm{clean}} \mid x, \sigma].
\end{equation}
This objective encourages the model to preserve statistically predictable and reproducible components at each noise scale while suppressing unpredictable perturbations, aligning with the goal of maintaining stable physiological structures under varying noise conditions.

When $\sigma$ spans a wide range, the magnitude and gradient distributions of inputs differ significantly across noise levels, making direct learning prone to scale inconsistency. To alleviate this issue, we adopt the EDM preconditioning formulation:
\begin{equation}
D_{\theta}(x, \sigma)
= c_{\mathrm{skip}}(\sigma) x
+ c_{\mathrm{out}}(\sigma)
F_{\theta}\!\left(
c_{\mathrm{in}}(\sigma) x,\,
c_{\mathrm{noise}}(\sigma)
\right),
\end{equation}
where $F_{\theta}(\cdot)$ denotes the learnable backbone network. The scale-dependent coefficients map inputs at different noise levels into a unified normalized space, allowing the backbone to focus on learning residuals relevant to structure recovery, while the skip connection helps preserve waveform morphology and phase consistency in low-noise regimes.

During training, only the single-step denoising mapping is learned. At inference time, these local denoising behaviors are composed along a monotonically decreasing noise schedule using a deterministic solver (\texttt{solve++}) [31], forming a stable reverse trajectory from high noise to low noise. This inference strategy emphasizes structural consistency and reproducibility rather than sample diversity.

Under this unified diffusion perspective, the same modeling formulation can be instantiated at different levels of the system. In the following, we describe its concrete realizations in the data domain and the feature domain, respectively.

\subsection{Data-Domain Stabilization}
\label{sec:data_domain}
\begin{figure}[!t]
    \centering
    \includegraphics[width=0.95\columnwidth]{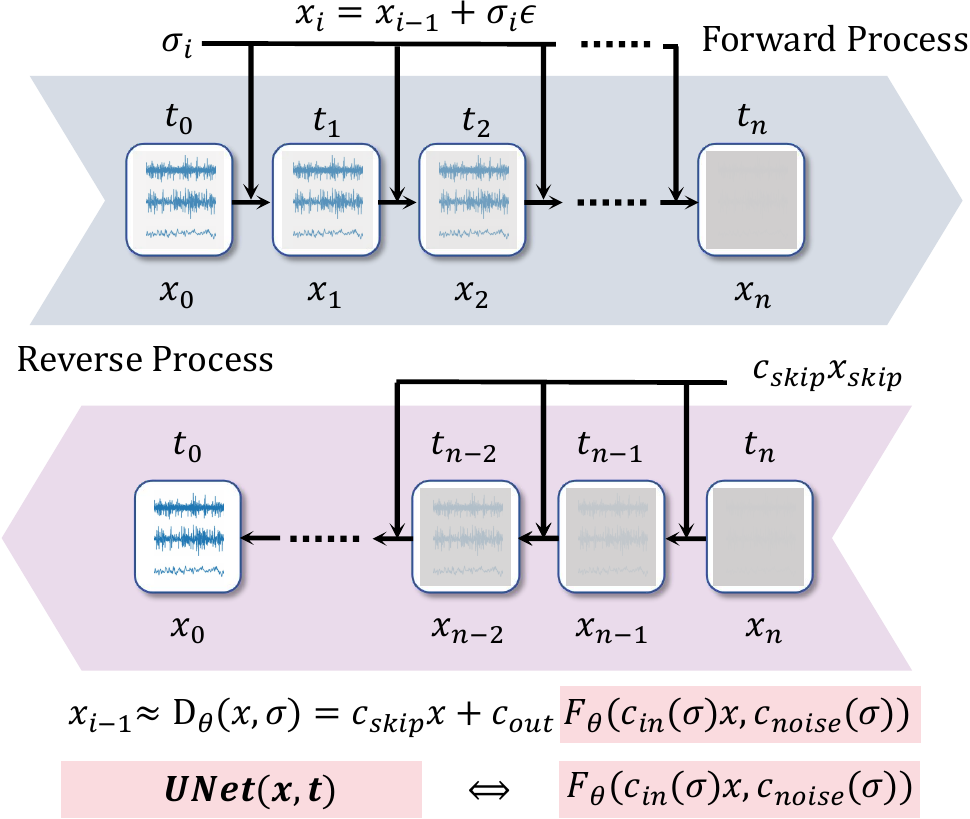}
    \caption{The forward process adds noise to PSG signals along predefined scales, and the reverse process uses a U-Net–based denoising function conditioned on the noise level to reconstruct clean and physiologically consistent signals [32].}
    \label{fig:single}
\end{figure}

In the data domain, the diffusion framework is instantiated as a continuous-scale stabilization module that directly operates on raw one-dimensional PSG waveforms. The goal of this module is not signal generation, but robust recovery of physiologically meaningful temporal structures from corrupted or noisy observations, thereby providing a stable input foundation for downstream modeling.

\paragraph{Continuous-scale perturbation and denoising formulation.}
Let $x_{\text{clean}}$ denote waveform segments that preserve stable physiological structures. To model the continuously varying noise conditions encountered in real-world PSG recordings, training samples are constructed by applying additive perturbations with continuously sampled noise scales:
\begin{equation}
x = x_{\text{clean}} + \sigma \varepsilon,\quad 
\varepsilon \sim \mathcal{N}(0, I),
\tag{6}
\end{equation}
where $\sigma$ controls the corruption strength. The denoising objective is to recover the underlying stable waveform conditioned on both the corrupted input and the noise scale:
\begin{equation}
x_{\text{clean}} \approx D_\theta(x, \sigma).
\tag{7}
\end{equation}
This formulation allows the model to learn a unified denoising operator that adapts smoothly across different noise regimes, rather than relying on discrete noise levels or stage-specific models.

\paragraph{Architecture and scale-adaptive behavior.}
The denoiser $D_\theta$ adopts an EDM-preconditioned one-dimensional U-Net architecture with a multi-scale encoder--decoder topology and skip connections. Residual modeling is incorporated within each scale to stabilize optimization. From a signal processing perspective, this architecture can be interpreted as a family of nonlinear filters with shared parameters, whose effective behavior varies continuously with the noise scale.

Specifically, in low-noise regimes, the skip connections dominate the signal path, ensuring temporal continuity and phase preservation; in higher-noise regimes, the residual pathways become more influential, focusing on reconstructing structural components that are partially obscured by noise.

\paragraph{Deterministic inference and system integration.}
At inference time, data-domain stabilization is performed deterministically using the \texttt{solve++} procedure, which composes the learned single-step denoising operators along a monotonically decreasing noise schedule. This deterministic formulation suppresses stochastic variations in low-noise regimes, effectively reducing waveform jitter and structural drift---an important property for preserving slow-wave morphology and spindle patterns critical to sleep staging.

The data-domain diffusion module is trained independently and frozen after convergence. Its output serves as a front-end enhancement interface and is subsequently transformed into time--frequency representations for downstream processing. No classification loss is back-propagated through this module, allowing it to focus exclusively on recovering stable physiological structures.

\subsection{Feature-Domain Stabilization}
\label{sec:feature_domain}
\begin{figure}[!t]
    \centering
    \includegraphics[width=0.95\columnwidth]{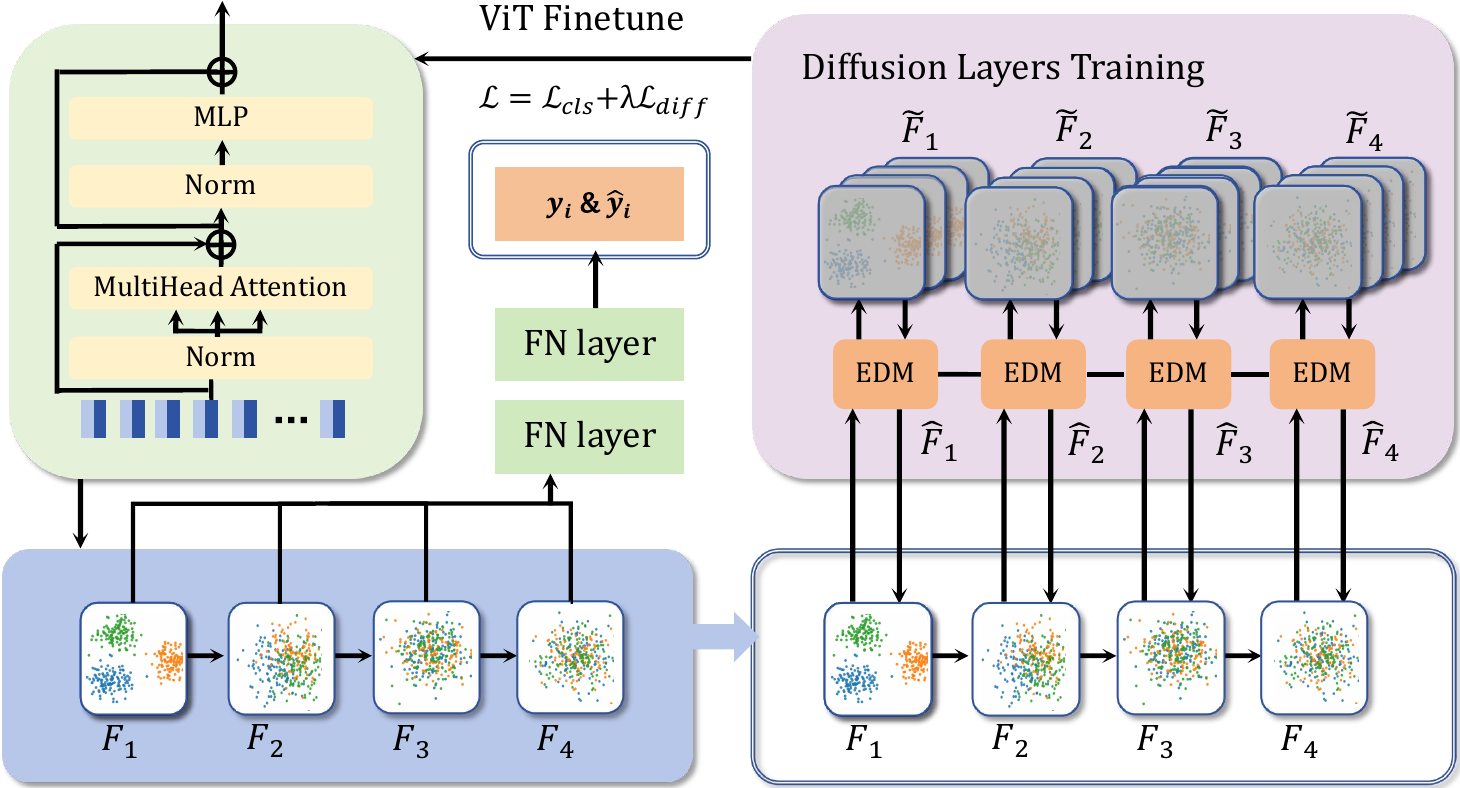}
    \caption{Multi-level features extracted by the Vision Transformer are refined through feature-domain diffusion modules (EDM), where diffusion-based correction is applied at each feature level to reduce representation instability. The stabilized features are then fed into fusion layers for supervised finetuning and final sleep stage classification.}
    \label{fig:feature_domain}
\end{figure}

While data-domain stabilization improves input signal quality, performance degradation often persists in cross-subject and cross-dataset settings. This observation suggests that instability arises not only from waveform-level noise, but also from drift and dispersion in high-dimensional feature representations learned by deep models. In standard discriminative training, separability is primarily enforced at the classifier output, whereas intermediate representations are not explicitly constrained to be stable or reproducible.

To address this limitation, we introduce a diffusion-based stabilization mechanism in the feature domain. Unlike generative diffusion models, the objective here is feature correction rather than feature synthesis: given potentially unstable student features, the model learns to project them back toward stable reference regions in the representation space.

\paragraph{Teacher--student reference mechanism.}
In the feature domain, there is no physically meaningful notion of a ``clean'' target. Directly using student features as regression targets would result in self-referential supervision and amplify existing drift. To provide a stable reference, we introduce an exponential moving average (EMA) teacher network.

Given an input $X$, the student backbone produces multi-level feature representations:
\begin{equation}
F_k = f_\theta^{(k)}(X), \quad k \in \{1,\dots,K\},
\tag{8}
\end{equation}
and the corresponding teacher features are defined as
\begin{equation}
F_k^{t} = f_{\bar{\theta}}^{(k)}(X), \quad
\bar{\theta} \leftarrow \mu \bar{\theta} + (1 - \mu)\theta.
\tag{9}
\end{equation}
The EMA teacher evolves slowly over training iterations and defines a smooth, stable reference manifold in the representation space, which serves as an anchor for feature stabilization.

\paragraph{Noise injection and feature correction via diffusion.}
Around each teacher feature $F_k^{t}$, local neighborhood samples are constructed by applying continuous-scale perturbations:
\begin{equation}
\tilde{F}_k = F_k^{t} + \sigma \varepsilon, \quad 
\varepsilon \sim \mathcal{N}(0, I).
\tag{10}
\end{equation}
A feature-domain diffusion module $D_{\psi}^{k}$ is then trained to map perturbed features back toward the reference center:
\begin{equation}
\hat{F}_k = D_{\psi}^{k}(\tilde{F}_k, \sigma).
\tag{11}
\end{equation}
The corresponding feature-domain diffusion loss is defined as
\begin{equation}
L_{\text{feat}}^{(k)} =
\mathbb{E}\!\left[
w(\sigma)\,
\big\|
\hat{F}_k - F_k^{t}
\big\|_2^2
\right],
\tag{12}
\end{equation}
which learns a scale-conditioned local correction operator that contracts features toward stable regions and suppresses unstable directions in the representation space.

\paragraph{Training strategy and inference behavior.}
Training follows a two-stage strategy. In the first stage, the backbone network is frozen and only the feature-domain diffusion modules are optimized, enabling them to learn stable correction mappings in the vicinity of the teacher-defined manifold. In the second stage, the stabilized features are forwarded to the fusion module and classification head, and the entire system is optimized under a joint objective:
\begin{equation}
L_{\text{total}} = L_{\text{cls}} + 
\lambda \sum_{k \in K} L_{\text{feat}}^{(k)},
\tag{13}
\end{equation}
where $L_{\text{cls}}$ denotes the standard cross-entropy loss.

At inference time, neither the teacher network nor explicit noise sampling is required. The trained feature-domain diffusion modules are applied deterministically to the student features at an infinitesimal noise scale ($\sigma \to 0$), acting as lightweight projection operators that improve feature consistency without increasing inference complexity.

By applying stabilization at multiple network depths, the proposed framework explicitly accounts for the hierarchical nature of sleep staging cues and effectively mitigates feature drift across layers under varying recording conditions.

\section{Experiment}
\renewcommand{\thesubsection}{\thesection.\arabic{subsection}}  
\setcounter{subsection}{0} 
\subsection{Experiment Setting}

\begin{table*}[t]
\centering
\caption{Statistics of Sleep Stage Distribution in Different Datasets}
\label{tab:1}
\small
\setlength{\tabcolsep}{9pt}   
\begin{tabular}{@{}l c c c | c c c c c | r @{}}
\toprule
\multirow{2}{*}{\textbf{Dataset}} 
 & \multirow{2}{*}{\textbf{Subjects}} 
 & \multirow{2}{*}{\textbf{Channels}} 
 & \multirow{2}{*}{\textbf{Freq.}} 
 & \multicolumn{5}{c|}{\textbf{Epoch numbers}} 
 & \multirow{2}{*}{\textbf{Total}} \\
\cmidrule(lr{0.5em}){5-9}
 &  &  &  & Wake    & N1     & N2      & N3     & REM    &  \\
\midrule
Sleepedf-20 & 20  & 3 & 100 Hz & 8285  & 2804  & 17799 & 5703 & 7717 & 42308 \\
            &     &        &        & 19.6\% & 6.6\% & 42.1\% & 13.5\% & 18.2\% &       \\
\midrule
Sleepedf-78 & 78  & 3 & 100 Hz & 63802 & 21229 & 68645 & 12883 & 25651 & 192210 \\
            &     &        &        & 33.2\% & 11.0\% & 35.71\% & 6.7\% & 13.34\% &       \\
\midrule
SHHS        & 329 & 3  & 125 Hz & 46369 & 10304 & 142125 & 60153 & 65953 & 324854 \\
            &     &        &        & 14.3\% & 3.2\% & 43.7\% & 18.5\% & 20.3\% &       \\
\midrule
ISRUC-S3    & 10  & 3  & 200 Hz & 1651  & 1215  & 2609  & 2014  & 1060  & 8549  \\
            &     &        &        & 20.5\% & 14.2\% & 30.5\% & 23.6\% & 12.4\% &       \\

\bottomrule
\setlength{\textfloatsep}{6pt}
\setlength{\intextsep}{6pt}
\end{tabular}
\end{table*}

\begin{table}[t]
\setlength{\tabcolsep}{30pt}
\centering
\caption{Key Hyperparameters Used in Experiments}
\label{tab:2}
\begin{tabular}{c c}
\hline
Symbol & Value \\
\hline
$[\sigma^{data}_{min}, \sigma^{data}_{max}]$ & [0.05, 0.2] \\
$\sigma_{data}$ & 0.5 \\
$N_{data}$ & 8 / 12 / 16 \\
$\sigma^{feat}_{max}$ & 0.2 \\
$K$ & 4 \\
$\lambda$ & 0.2 \\
$\mu$ & 0.999 \\
ViT dim & 256 \\
ViT depth & 4 \\
ViT heads & 4 \\
Optimizer & AdamW \\
Learning rate & $1\times10^{-4}$ \\
\hline
\end{tabular}
\end{table}

DSSNet is evaluated on multiple publicly available sleep staging datasets that exhibit substantial diversity in terms of subject population, sampling frequency, and sleep stage distribution. The basic statistics and stage distributions of these datasets are summarized in Table~\ref{tab:1}. Unless otherwise specified, all experiments are conducted under a unified experimental protocol, where the model architecture parameters and training-related hyperparameters are fixed to a default configuration, with their specific values reported in Table~\ref{tab:2}. To ensure fair and reproducible comparisons, all competing methods strictly follow the parameter settings reported in their original papers or official implementations.

\subsection{Evaluation Metrics}
\label{sec:metrics}

To comprehensively evaluate the classification performance and sequence-level consistency of the proposed framework, we adopt a set of widely used evaluation metrics, including overall accuracy (ACC), macro-averaged precision (MPR), macro-averaged recall (MRE), macro-averaged F1-score (Macro F1), and Cohen’s kappa coefficient ($\kappa$).

Overall accuracy is defined as
\begin{equation}
\mathrm{ACC} = \frac{TP + TN}{TP + TN + FP + FN},
\label{eq:acc}
\end{equation}
where $TP$, $TN$, $FP$, and $FN$ denote the numbers of true positives, true negatives, false positives, and false negatives, respectively.

For each sleep stage $k$, the per-class precision and recall are computed as
\begin{equation}
\mathrm{PR}_k = \frac{TP_k}{TP_k + FP_k}, \quad
\mathrm{RE}_k = \frac{TP_k}{TP_k + FN_k}.
\label{eq:pr_re}
\end{equation}

The macro-averaged precision and recall are obtained by averaging over all classes:
\begin{equation}
\mathrm{MPR} = \frac{1}{N_{\text{classes}}} \sum_{k=1}^{N_{\text{classes}}} \mathrm{PR}_k,
\label{eq:mpr}
\end{equation}
\begin{equation}
\mathrm{MRE} = \frac{1}{N_{\text{classes}}} \sum_{k=1}^{N_{\text{classes}}} \mathrm{RE}_k.
\label{eq:mre}
\end{equation}

The macro-averaged F1-score (Macro F1) is defined as the harmonic mean of macro-averaged precision and recall:
\begin{equation}
\mathrm{Macro\ F1} = \frac{2 \times \mathrm{MPR} \times \mathrm{MRE}}{\mathrm{MPR} + \mathrm{MRE}}.
\label{eq:macro_f1}
\end{equation}

Cohen’s kappa coefficient ($\kappa$) measures the agreement between predicted and true labels beyond chance and is particularly suitable for imbalanced classification tasks:
\begin{equation}
\kappa = \frac{\mathrm{ACC} - p_e}{1 - p_e},
\label{eq:kappa}
\end{equation}
where $\mathrm{ACC}$ is defined in Eq.~(\ref{eq:acc}), and $p_e$ denotes the probability of agreement by chance, computed as
\begin{equation}
p_e = \frac{1}{N^2} \sum_{k=1}^{N_{\text{classes}}} n_{k1} \, n_{k2}.
\label{eq:pe}
\end{equation}

Here, $N$ denotes the total number of samples, $n_{k1}$ is the number of samples whose ground-truth label belongs to class $k$, and $n_{k2}$ is the number of samples predicted as class $k$.

\begin{table*}[!t]
\centering
\caption{Comparison with State-of-the-Art Methods on Five Public Datasets}
\label{tab.3}
\small
\setlength{\aboverulesep}{0.0ex}
\setlength{\belowrulesep}{0.1ex}
\setlength{\tabcolsep}{4.0pt}
\renewcommand{\arraystretch}{1.05}

\resizebox{\textwidth}{!}{%
\begin{tabular}{@{}l l ccc ccccc c@{}}
\toprule
\textbf{Dataset} & \textbf{Method} & \textbf{Architecture} & \textbf{Acc} & \textbf{MF1} & \textbf{Kappa} & \textbf{W} & \textbf{N1} & \textbf{N2} & \textbf{N3} & \textbf{R} \\
\midrule

\multirow{9}{*}{ISRUC-S3}
 & MVF-SleepNet [21]      & CNN + LSTM         & 84.1 & 82.8 & 0.795 & 90.0 & 62.5 & 83.3 & 91.1 & 87.3 \\
 & SVM [14]               & Traditional        & 71.4 & 67.2 & 0.626 & 82.4 & 42.8 & 72.4 & 81.5 & 56.9 \\
 & RF [15]                & Traditional        & 70.2 & 68.5 & 0.616 & 83.8 & 47.0 & 67.1 & 76.3 & 68.4 \\
 & MixSleepNet [33]       & CNN + Transformer  & 83.0 & 82.1 & 0.782 & 89.9 & 62.5 & 81.9 & 89.9 & 86.0 \\
 & XSleepNet1 [12]        & CNN + LSTM         & 82.5 & 80.8 & 0.774 & 90.1 & 58.6 & 82.5 & 88.7 & 84.3 \\
 & XSleepNet2 [12]         & CNN + LSTM         & 82.6 & 81.0 & 0.774 & 89.9 & 59.0 & 82.6 & 88.4 & 84.9 \\
 & SeqSleepNet  [20]      & LSTM               & 78.9 & 76.3 & 0.730 & 83.6 & 43.9 & 79.3 & 87.9 & 86.7 \\
 & DSSNet(ours)*              & Diffusion                 & 85.0 & 83.3 & 0.807 & 90.3 & 61.6 & 86.9 & 88.8 & 88.9 \\
 & \textbf{DSSNet(ours)}      & \textbf{Diffusion}        & \textbf{86.7} & \textbf{84.9} & \textbf{0.829} & \textbf{90.3} & \textbf{64.5} & \textbf{89.9} & \textbf{89.3} & \textbf{90.4} \\
\midrule

\multirow{11}{*}{SHHS}
 & XSleepNet1 [12]        & CNN + LSTM         & 87.5 & 81.0 & 0.826 & 91.6 & 51.4 & 88.5 & 85.0 & 88.4 \\
 & XSleepNet2 [12]         & CNN + LSTM         & 87.6 & 80.7 & 0.826 & 92.0 & 49.9 & 88.3 & 85.0 & 88.2 \\
 & DilatedSleepNet [19]   & CNN                & 85.4 & 78.7 & 0.800 & 86.1 & 47.7 & 87.1 & 84.7 & 87.8 \\
 & AttnSleep [18]       & CNN                & 84.2 & 75.3 & 0.780 & 86.7 & 33.2 & 87.1 & 87.1 & 82.1 \\
 & DeepSleepNet [16]     & CNN                & 81.0 & 73.9 & 0.730 & 85.4 & 40.5 & 82.5 & 79.3 & 81.9 \\
 & SleepEEGNet [13]       & CNN                & 73.9 & 68.4 & 0.650 & 81.3 & 34.4 & 73.4 & 75.9 & 77.0 \\
 & SleepTransformer [34]   & Transformer         & 87.7 & 80.1 & 0.828 & 92.2 & 46.1 & 88.3 & 85.2 & 88.6 \\
 & FlexibleSleepNet [37]   & CNN                & 87.6 & 79.5 & 0.830 & 92.3 & 40.0 & 88.8 & 87.0 & 89.7 \\
 & SleepViTransformer [35] & Transformer         & 88.1 & 79.8 & 0.830 & 93.4 & 44.4 & 88.5 & 88.5 & 88.3 \\
 & DSSNet(ours)*              & Diffusion                 & 88.5 & 82.6 & 0.839 & 90.5 & 54.2 & 89.7 & 88.8 & 89.8 \\
 & \textbf{DSSNet(ours)}      & \textbf{Diffusion}        & \textbf{89.7} & \textbf{84.0} & \textbf{0.855} & \textbf{91.6} & \textbf{56.9} & \textbf{90.8} & \textbf{89.6} & \textbf{91.4} \\
\midrule

\multirow{12}{*}{SleepEDF-20}
 & XSleepNet1 [12]         & CNN + LSTM         & 86.0 & 80.0 & 0.810 & 91.3 & 49.5 & 88.0 & 86.9 & 84.2 \\
 & XSleepNet2 [12]         & CNN + LSTM         & 86.3 & 80.6 & 0.813 & 92.2 & 51.8 & 88.0 & 86.8 & 83.9 \\
 & SailentSleepNet [36]   & CNN-Transformer    & 87.5 & 83.0 & --    & 92.3 & 56.2 & 89.9 & 87.2 & 89.2 \\
 & SeqSleepNet [20]       & LSTM               & 86.0 & 79.7 & 0.810 & --   & --   & --   & --   & --   \\
 & DilatedSleepNet [19]   & CNN                & 86.8 & 81.9 & 0.820 & 90.8 & 53.3 & 89.4 & 89.9 & 85.8 \\
 & AttnSleep [18]         & CNN                & 84.4 & 78.1 & 0.790 & 89.7 & 42.6 & 88.8 & 90.2 & 79.0 \\
 & DeepSleepNet [16]      & CNN                & 82.0 & 76.9 & 0.760 & 84.7 & 46.6 & 85.9 & 84.8 & 82.4 \\
 & SleepEEGNet [13]       & CNN                & 84.3 & 79.7 & 0.790 & 89.2 & 52.2 & 86.8 & 85.1 & 85.0 \\
 & FlexibleSleepNet [37]  & CNN                & 86.9 & 81.9 & 0.824 & 92.8 & 57.2 & 89.8 & 85.0 & 87.2 \\
 & TinySleepNet [17]      & CNN                & 85.4 & 80.5 & --    & 90.1 & 51.4 & 88.5 & 88.3 & 84.3 \\
 & SleepViTransformer [35] & Transformer         & 87.8 & 81.5 & 0.834 & 93.8 & 48.4 & 89.2 & 88.4 & 87.9 \\
 & \textbf{DSSNet(ours)}      & \textbf{Diffusion} & \textbf{89.2} & \textbf{85.5} & \textbf{0.852} & \textbf{94.0} & \textbf{62.3} & \textbf{91.7} & \textbf{88.0} & \textbf{89.2} \\
\midrule

\multirow{12}{*}{SleepEDF-78}
 & SailentSleepNet [36]   & CNN-Transformer    & 84.1 & 79.5 & --    & 93.3 & 54.2 & 85.8 & 78.3 & 85.8 \\
 & SeqSleepNet [20]       & LSTM               & 83.8 & 78.2 & 0.780 & --   & --   & --   & --   & --   \\
 & DilatedSleepNet [19]   & CNN                & 83.2 & 77.5 & 0.770 & 92.7 & 47.2 & 85.5 & 81.8 & 80.5 \\
 & AttnSleep [18]         & CNN                & 81.3 & 75.1 & 0.740 & 92.0 & 42.0 & 85.0 & 82.1 & 74.2 \\
 & DeepSleepNet [16]      & CNN                & 76.9 & 70.7 & 0.690 & 90.8 & 44.8 & 78.5 & 67.9 & 71.3 \\
 & SleepEEGNet [13]       & CNN                & 80.0 & 73.6 & 0.730 & 91.7 & 44.1 & 82.5 & 73.5 & 75.2 \\
 & SleepTransformer [34]  & Transformer         & 81.4 & 74.3 & 0.743 & 91.7 & 40.4 & 84.3 & 77.9 & 77.2 \\
 & FlexibleSleepNet [37]  & CNN                & 87.0 & 82.7 & 0.820 & 95.3 & 59.9 & 88.0 & 84.4 & 86.1 \\
 & TinySleepNet [17]      & CNN                & 83.1 & 78.1 & --    & 92.8 & 51.0 & 85.3 & 81.1 & 80.3 \\
 & SleepViTransformer [35]& Transformer         & 85.0 & 79.1 & 0.792 & 93.6 & 49.4 & 86.4 & 79.3 & 86.9 \\
 & FSDM [38]              & Diffusion          & 82.6 & 0.753 & 0.757 & --   & --   & --   & --   & --   \\
 & \textbf{DSSNet(ours)}      & \textbf{Diffusion} & \textbf{88.0} & \textbf{84.2} & \textbf{0.836} & \textbf{94.3} & \textbf{62.3} & \textbf{90.3} & \textbf{84.6} & \textbf{89.2} \\
\bottomrule
\end{tabular}%
}
\end{table*}

\subsection{Comparison with State-of-the-Art Methods}

To systematically evaluate the performance and generalization capability of DSSNet(ours), we compare it with a wide range of representative sleep staging methods. The comparison results are summarized in Table~\ref{tab.3}. In the table, results marked with ``*'' correspond to a cross-dataset evaluation setting, where the model is trained only on the SleepEDF-78 dataset and directly tested on the target dataset.

On the ISRUC-S3 dataset, DSSNet achieves the best overall performance under the same-dataset training setting. Compared with the strongest competing method, MVF-SleepNet, DSSNet improves ACC, MF1, and $\kappa$ by 2.6\%, 2.1\%, and 0.034, respectively. From a stage-wise perspective, DSSNet attains an accuracy of 89.9\% on the N2 stage, outperforming MVF-SleepNet by 6.6\%, and also achieves a 3.1\% improvement on the REM stage. Under the cross-dataset evaluation setting, DSSNet still achieves an ACC of 85.0\% and a $\kappa$ of 0.807. Its overall performance not only significantly surpasses traditional methods, but also exceeds several deep models trained directly on the target dataset, demonstrating strong cross-dataset generalization capability.

On the SHHS dataset, DSSNet also delivers the best performance across overall metrics. Compared with the closest Transformer-based competitor, SleepViTransformer, DSSNet improves MF1 from 79.8\% to 84.0\% and $\kappa$ from 0.830 to 0.855, while achieving a 2.1\% ACC gain over FlexibleSleepNet. Stage-wise results show that DSSNet reaches an accuracy of 56.9\% on the N1 stage, outperforming SleepViTransformer by 12.5\%, and achieves 91.4\% accuracy on the REM stage, which is the highest reported on this dataset. In the cross-dataset setting, DSSNet maintains an ACC of 88.5\% and a $\kappa$ of 0.839, achieving performance comparable to or better than multiple methods trained on the same dataset, further indicating its robustness to data distribution shifts.

On the SleepEDF-20 dataset, DSSNet achieves the best results on all overall evaluation metrics under the same-dataset training condition. Compared with the previously best-performing method, SleepViTransformer, DSSNet improves ACC from 87.8\% to 89.2\%, MF1 from 81.5\% to 85.5\%, and $\kappa$ from 0.834 to 0.852. In terms of stage-wise performance, DSSNet reaches 62.3\% accuracy on the N1 stage, exceeding FlexibleSleepNet by 5.1\%, and achieves 91.7\% accuracy on the N2 stage, with an improvement of approximately 1.8\%. Under the cross-dataset evaluation setting, DSSNet still attains an ACC of 88.5\% and a $\kappa$ of 0.839, clearly outperforming most competing methods trained on the target dataset.

On the SleepEDF-78 dataset, DSSNet again achieves the best overall performance. Compared with the strongest competing method, FlexibleSleepNet, DSSNet improves MF1 from 82.7\% to 84.2\%, $\kappa$ from 0.820 to 0.836, and ACC by 1.0\%. Stage-wise results show that DSSNet achieves an accuracy of 62.3\% on the N1 stage, outperforming FlexibleSleepNet by 2.4\%, and reaches 90.3\% accuracy on the N2 stage, with a 2.3\% improvement. These results indicate that DSSNet maintains consistent advantages in large-scale, multi-subject scenarios.

\subsection{Per-Class Performance Statistics Across Datasets}
Table~\ref{tab.4} summarizes the per-sleep-stage classification results of the proposed model across multiple datasets, reporting Precision, Recall, F1-score, Specificity, and the corresponding number of samples for each stage. This table provides fine-grained numerical support for the overall performance comparison and stage-level analysis discussed in the main text.

From the per-class statistics, it can be observed that classification performance varies across different sleep stages and datasets. In particular, the Wake, N2, N3, and REM stages exhibit relatively stable classification behavior across datasets, whereas the N1 stage consistently shows lower Precision and Recall. This reflects the inherent complexity of N1 in terms of physiological characteristics and transitional temporal patterns. These per-class results are consistent with the analysis in the main text regarding the varying classification difficulty across sleep stages.

\begin{table}[!t]
\centering
\caption{Per-Class Performance Across Datasets}
\label{tab.4}

\footnotesize
\setlength{\tabcolsep}{10pt}
\renewcommand{\arraystretch}{1.1}

\begin{tabular}{@{}l c c c c r@{}}
\toprule

\multicolumn{6}{c}{\textbf{SleepEDF-78}} \\
\midrule
\textbf{Stage} & \textbf{Pre.} & \textbf{Rec.} & \textbf{F1} & \textbf{Spec.} & \textbf{Samples} \\
\midrule
Wake & 0.9583 & 0.9278 & 0.9428 & 0.9800 & 63,802 \\
N1   & 0.6297 & 0.6215 & 0.6256 & 0.9546 & 21,229 \\
N2   & 0.8985 & 0.9083 & 0.9034 & 0.9430 & 68,645 \\
N3   & 0.8435 & 0.8476 & 0.8455 & 0.9887 & 12,883 \\
REM  & 0.8677 & 0.9185 & 0.8924 & 0.9784 & 25,651 \\
\midrule

\multicolumn{6}{c}{\textbf{SleepEDF-20}} \\
\midrule
\textbf{Stage} & \textbf{Pre.} & \textbf{Rec.} & \textbf{F1} & \textbf{Spec.} & \textbf{Samples} \\
\midrule
Wake & 0.9469 & 0.9340 & 0.9404 & 0.9872 & 8,285 \\
N1   & 0.5942 & 0.6605 & 0.6256 & 0.9680 & 2,804 \\
N2   & 0.9245 & 0.9097 & 0.9170 & 0.9460 & 17,799 \\
N3   & 0.8253 & 0.9432 & 0.8803 & 0.9689 & 5,703 \\
REM  & 0.9389 & 0.8501 & 0.8923 & 0.9877 & 7,717 \\
\midrule

\multicolumn{6}{c}{\textbf{SHHS}} \\
\midrule
\textbf{Stage} & \textbf{Pre.} & \textbf{Rec.} & \textbf{F1} & \textbf{Spec.} & \textbf{Samples} \\
\midrule
Wake & 0.9265 & 0.9059 & 0.9161 & 0.9880 & 46,319 \\
N1   & 0.5495 & 0.5893 & 0.5687 & 0.9842 & 10,304 \\
N2   & 0.9146 & 0.9011 & 0.9078 & 0.9346 & 142,125 \\
N3   & 0.8933 & 0.8989 & 0.8961 & 0.9756 & 60,153 \\
REM  & 0.9001 & 0.9274 & 0.9135 & 0.9738 & 65,953 \\
\midrule

\multicolumn{6}{c}{\textbf{EDF-ISRUC-S3}} \\
\midrule
\textbf{Stage} & \textbf{Pre.} & \textbf{Rec.} & \textbf{F1} & \textbf{Spec.} & \textbf{Samples} \\
\midrule
Wake & 0.8639 & 0.9461 & 0.9032 & 0.9643 & 1,651 \\
N1   & 0.6648 & 0.5745 & 0.6163 & 0.9520 & 1,215 \\
N2   & 0.8713 & 0.8666 & 0.8689 & 0.9438 & 2,609 \\
N3   & 0.8849 & 0.8918 & 0.8884 & 0.9665 & 1,914 \\
REM  & 0.8860 & 0.8914 & 0.8887 & 0.9820 & 1,160 \\
\midrule

\multicolumn{6}{c}{\textbf{EDF-SHHS}} \\
\midrule
\textbf{Stage} & \textbf{Pre.} & \textbf{Rec.} & \textbf{F1} & \textbf{Spec.} & \textbf{Samples} \\
\midrule
Wake & 0.9163 & 0.8945 & 0.9053 & 0.9864 & 46,319 \\
N1   & 0.5177 & 0.5676 & 0.5415 & 0.9827 & 10,304 \\
N2   & 0.9060 & 0.8886 & 0.8972 & 0.9283 & 142,125 \\
N3   & 0.8856 & 0.8903 & 0.8880 & 0.9739 & 60,153 \\
REM  & 0.8818 & 0.9153 & 0.8982 & 0.9687 & 65,953 \\
\midrule

\multicolumn{6}{c}{\textbf{ISRUC-S3}} \\
\midrule
\textbf{Stage} & \textbf{Pre.} & \textbf{Rec.} & \textbf{F1} & \textbf{Spec.} & \textbf{Samples} \\
\midrule
Wake & 0.8538 & 0.9588 & 0.9033 & 0.9607 & 1,651 \\
N1   & 0.7297 & 0.5778 & 0.6449 & 0.9645 & 1,215 \\
N2   & 0.8991 & 0.8984 & 0.8988 & 0.9557 & 2,609 \\
N3   & 0.8880 & 0.8986 & 0.8933 & 0.9673 & 1,914 \\
REM  & 0.8932 & 0.9155 & 0.9042 & 0.9828 & 1,160 \\

\bottomrule
\end{tabular}
\end{table}

\section{Visualization}
\begin{figure*}[!t]
\centering
\captionsetup{font=scriptsize}

\begin{subfigure}[t]{0.23\linewidth}
  \centering
  \includegraphics[width=\linewidth]{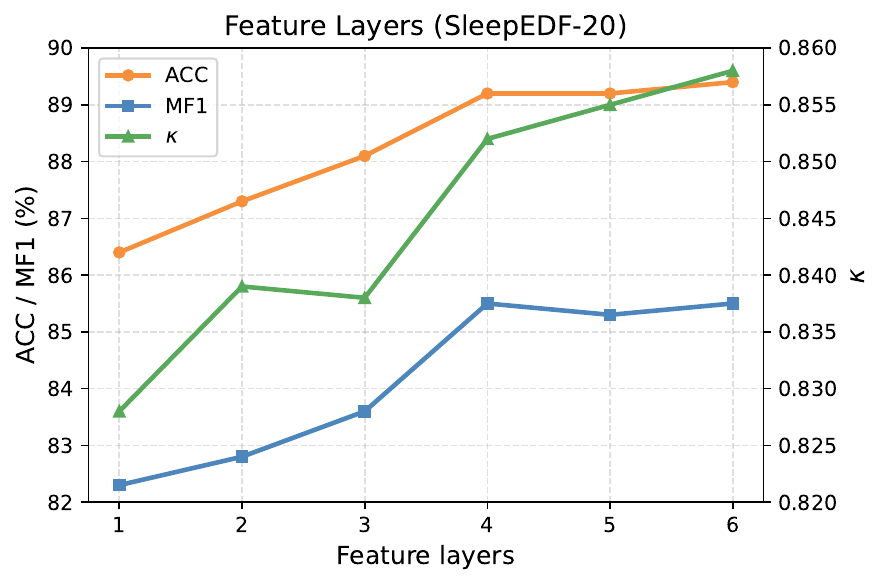}
  \caption{ISRUC-S3}
\end{subfigure}\hfill
\begin{subfigure}[t]{0.23\linewidth}
  \centering
  \includegraphics[width=\linewidth]{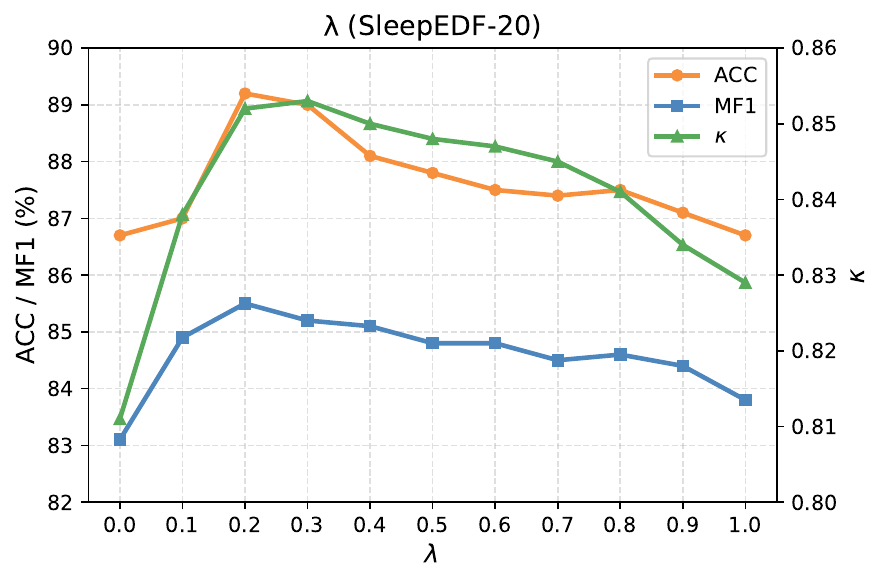}
  \caption{EDF-SHHS}
\end{subfigure}\hfill
\begin{subfigure}[t]{0.23\linewidth}
  \centering
  \includegraphics[width=\linewidth]{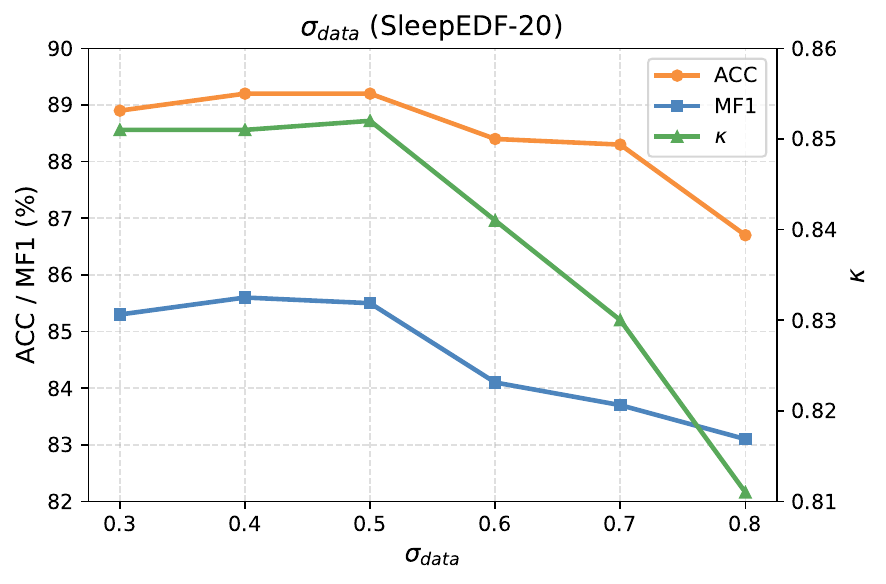}
  \caption{SleepEDF-20}
\end{subfigure}\hfill
\begin{subfigure}[t]{0.23\linewidth}
  \centering
  \includegraphics[width=\linewidth]{para_SleepEDF-20_sensi_sigma_data.pdf}
  \caption{SleepEDF-78}
\end{subfigure}

\par\vspace*{2pt}

\begin{subfigure}[t]{0.23\linewidth}
  \centering
  \includegraphics[width=\linewidth]{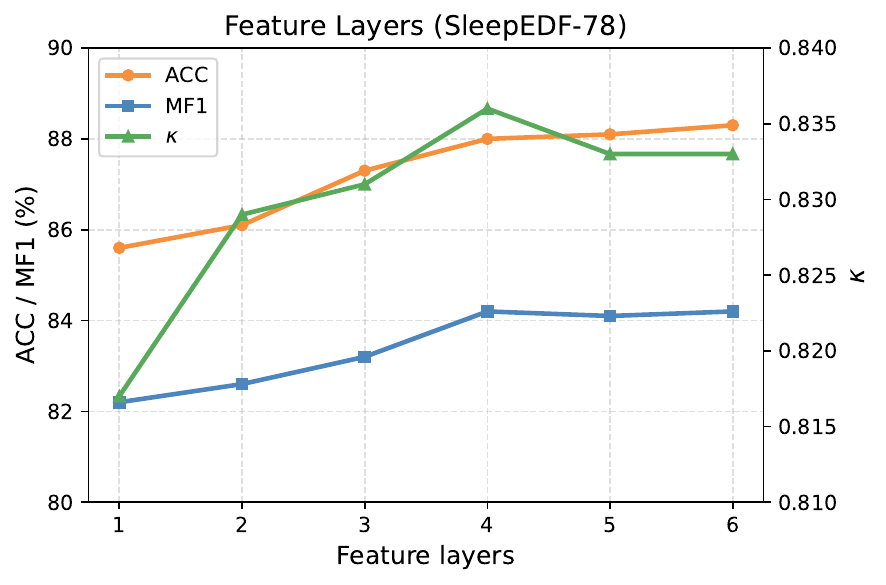}
  \caption{SHHS}
\end{subfigure}\hfill
\begin{subfigure}[t]{0.23\linewidth}
  \centering
  \includegraphics[width=\linewidth]{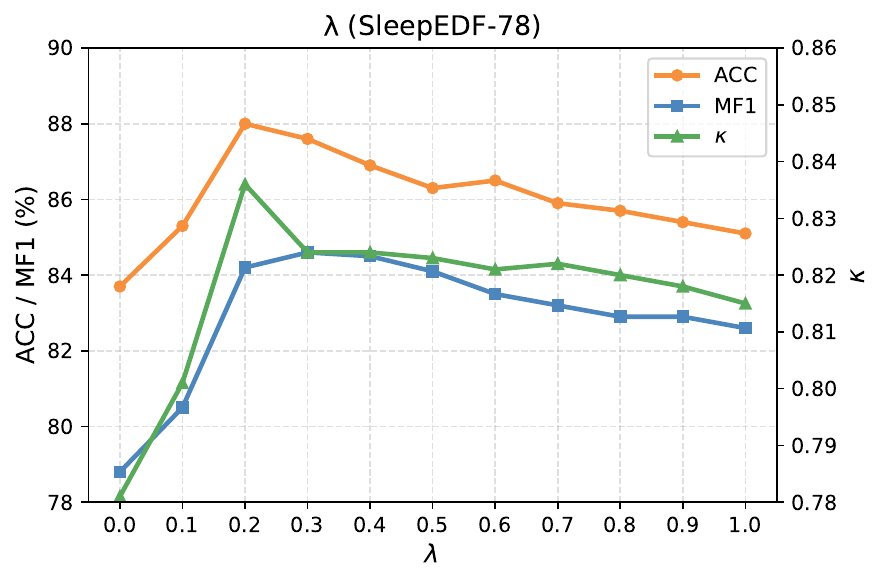}
  \caption{EDF-ISRUC-S3}
\end{subfigure}\hfill
\begin{subfigure}[t]{0.23\linewidth}
  \centering
  \includegraphics[width=\linewidth]{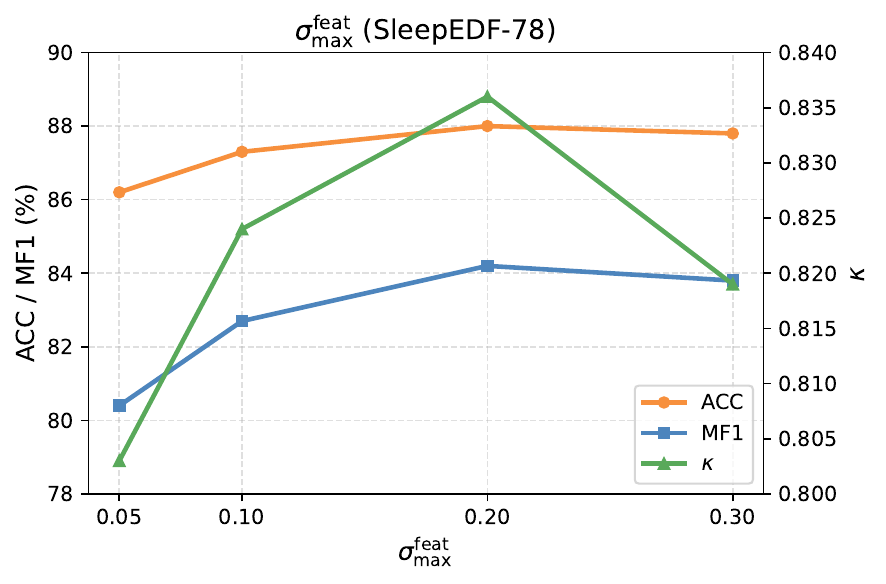}
  \caption{SleepEDF-78}
\end{subfigure}\hfill
\begin{subfigure}[t]{0.23\linewidth}
  \centering
  \includegraphics[width=\linewidth]{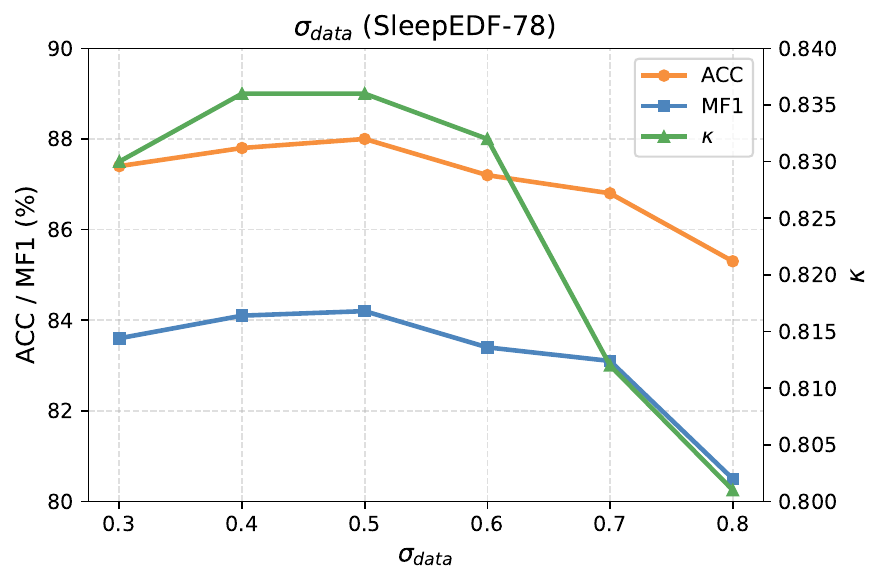}
  \caption{SleepEDF-78}
\end{subfigure}

\caption{Hyperparameter sensitivity analysis under different settings. Each curve represents performance variation with respect to one hyperparameter while keeping others fixed. The proposed model shows limited performance fluctuation, suggesting strong robustness.}
\label{fig.4}
\end{figure*}

\begin{figure*}[!t]
\centering
\captionsetup{font=scriptsize}

\begin{subfigure}[t]{0.30\linewidth}
  \centering
  \includegraphics[width=\linewidth]{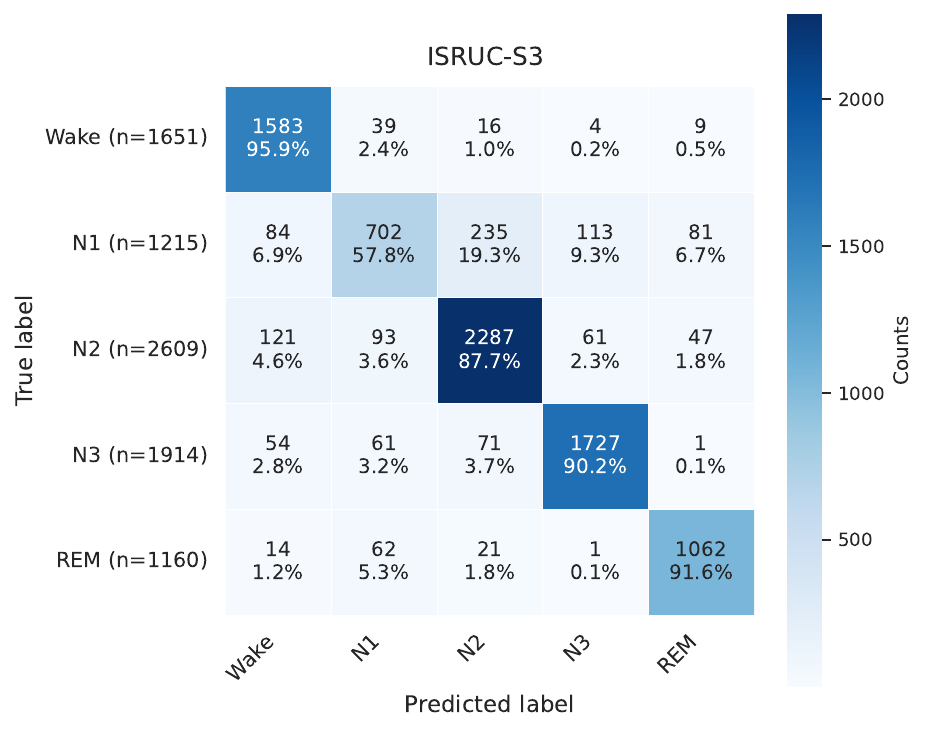}
  \caption{ISRUC-S3}
\end{subfigure}\hfill
\begin{subfigure}[t]{0.30\linewidth}
  \centering
  \includegraphics[width=\linewidth]{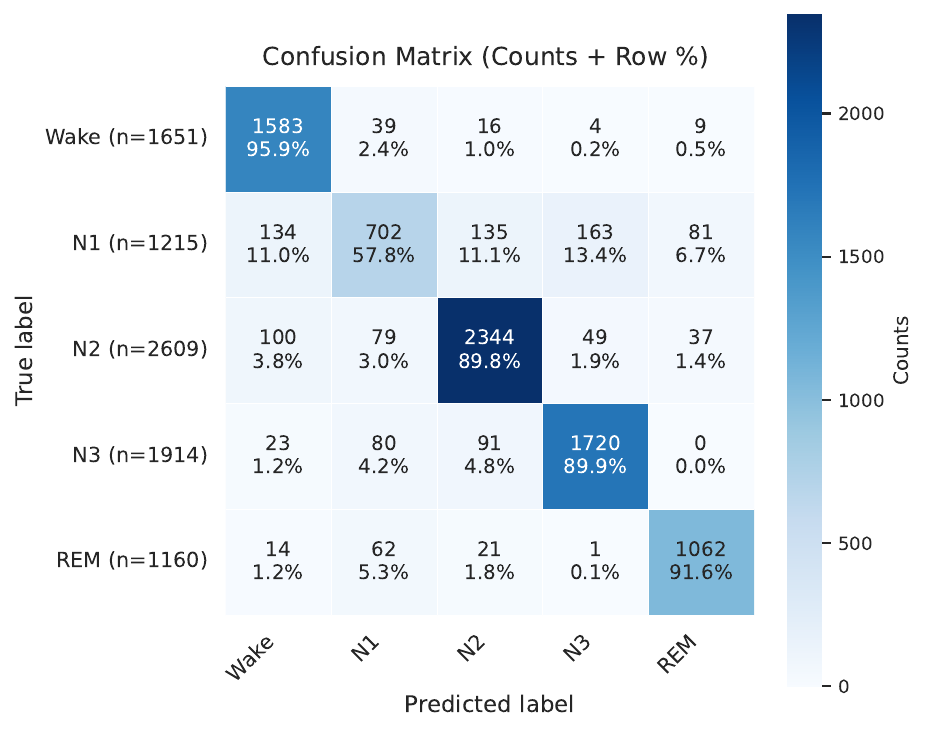}
  \caption{EDF-SHHS}
\end{subfigure}\hfill
\begin{subfigure}[t]{0.30\linewidth}
  \centering
  \includegraphics[width=\linewidth]{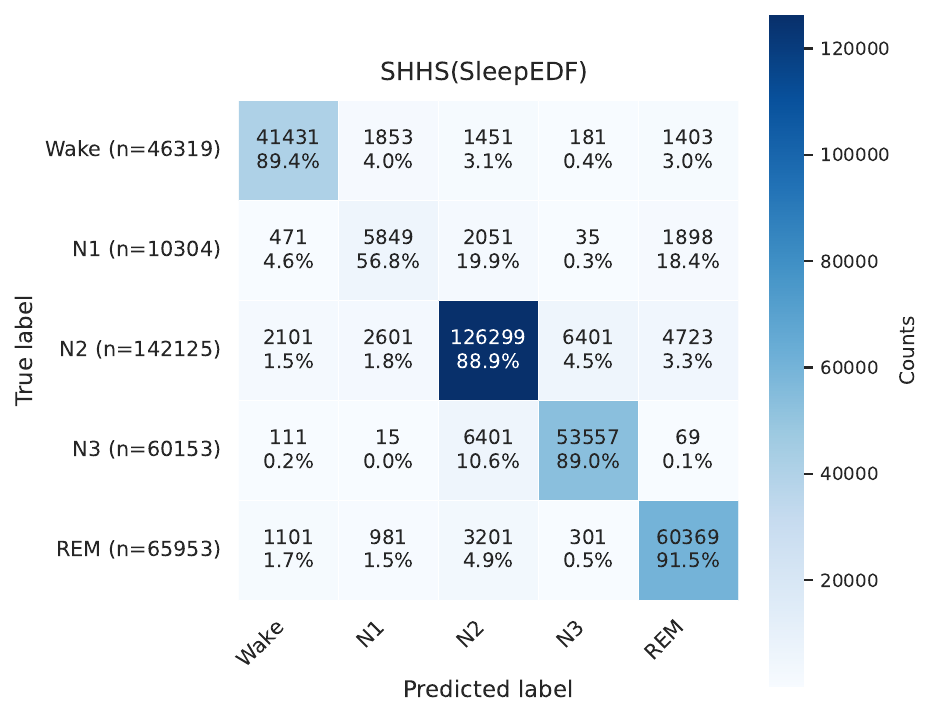}
  \caption{SleepEDF-20}
\end{subfigure}

\par\vspace*{2pt}

\begin{subfigure}[t]{0.30\linewidth}
  \centering
  \includegraphics[width=\linewidth]{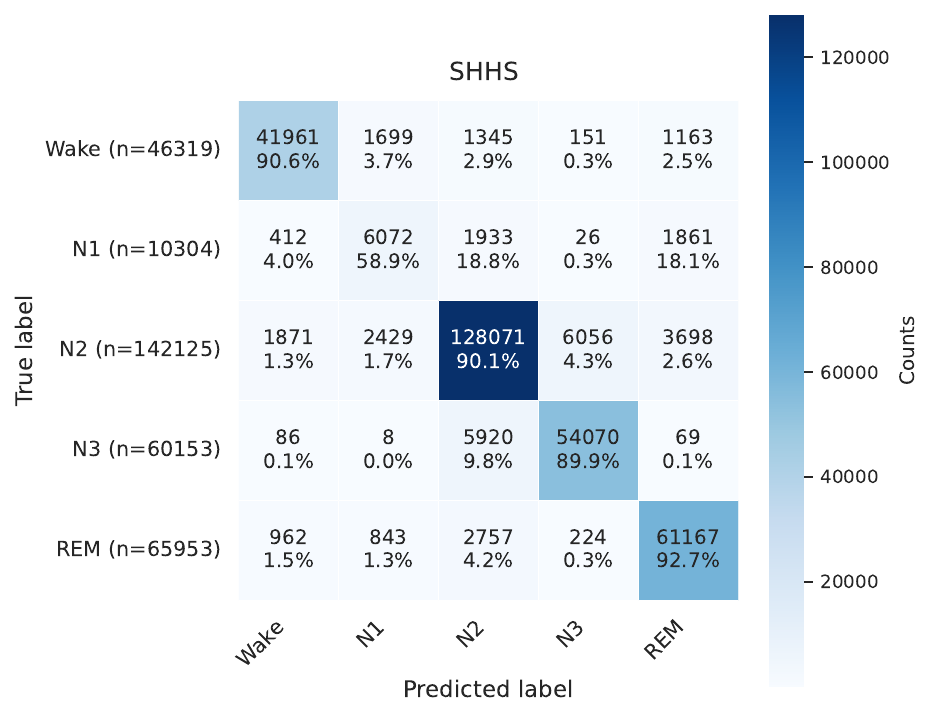}
  \caption{SHHS}
\end{subfigure}\hfill
\begin{subfigure}[t]{0.30\linewidth}
  \centering
  \includegraphics[width=\linewidth]{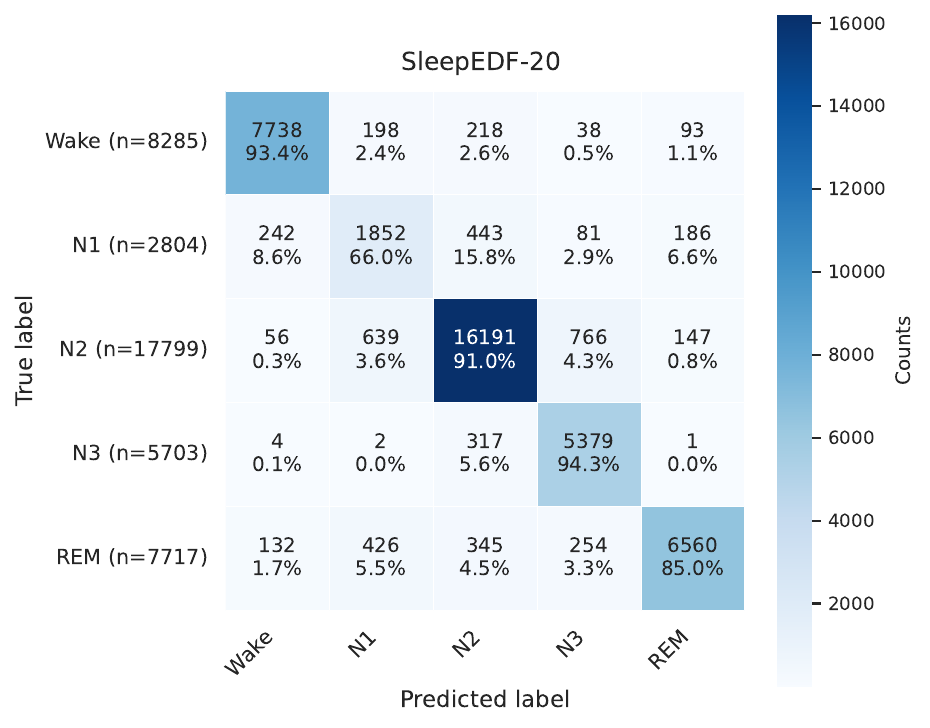}
  \caption{EDF-ISRUC-S3}
\end{subfigure}\hfill
\begin{subfigure}[t]{0.30\linewidth}
  \centering
  \includegraphics[width=\linewidth]{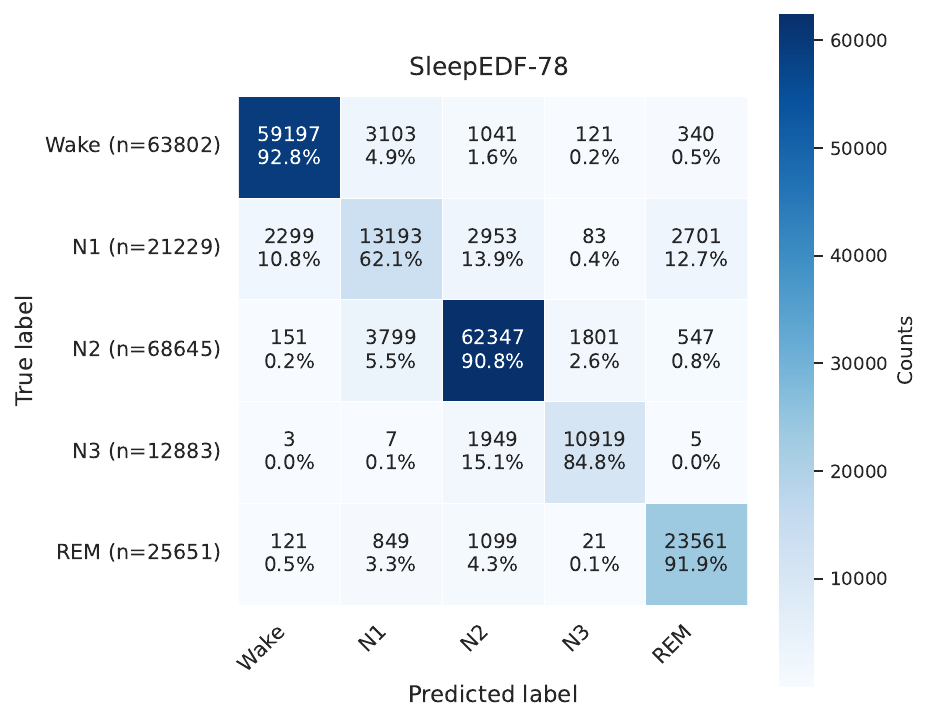}
  \caption{SleepEDF-78}
\end{subfigure}

\caption{Confusion matrices of sleep stage classification obtained from a single multi-channel model. Rows denote ground-truth labels and columns denote predicted labels. Subfigures (A–D) correspond to evaluations on four individual datasets, while (E–F) present cross-dataset results. The matrices indicate consistent classification behavior across datasets, with most confusions occurring between neighboring sleep stages.}
\label{fig.5}
\end{figure*}

\begin{figure*}[!t]
\centering
\captionsetup{font=scriptsize}

\begin{subfigure}[t]{0.30\linewidth}
  \centering
  \includegraphics[width=\linewidth]{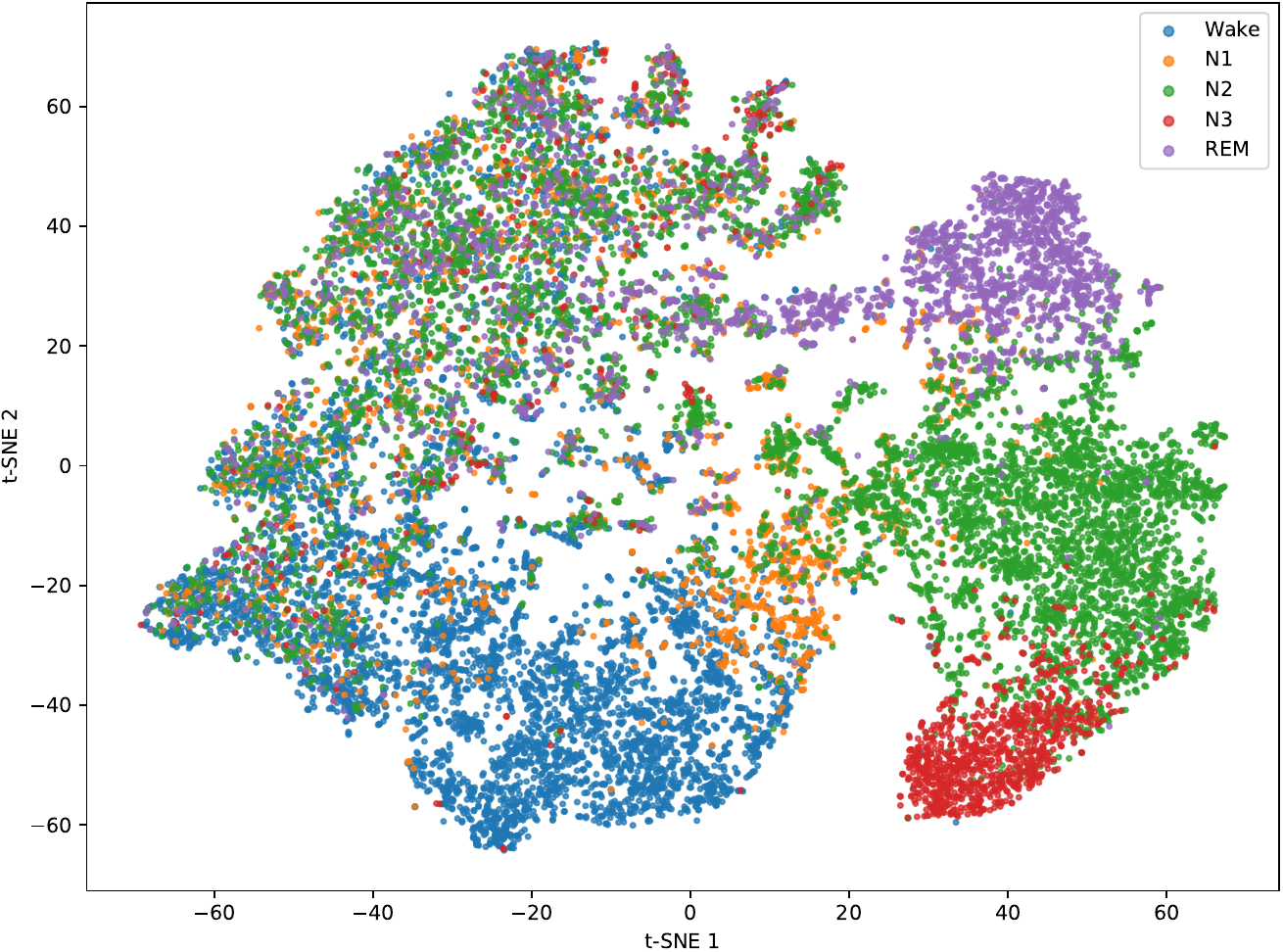}
  \caption{ISRUC-S3}
\end{subfigure}\hfill
\begin{subfigure}[t]{0.30\linewidth}
  \centering
  \includegraphics[width=\linewidth]{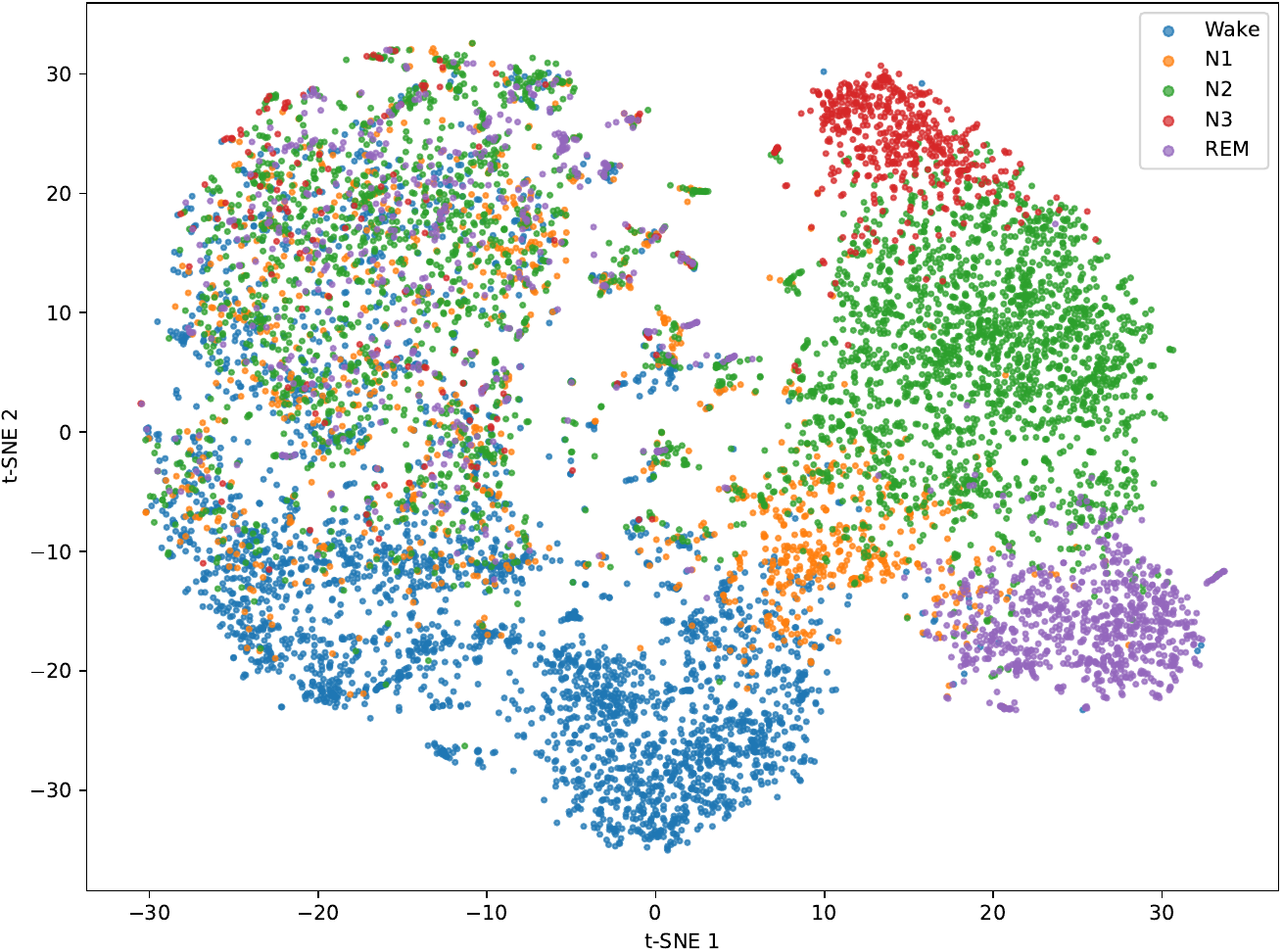}
  \caption{EDF-SHHS}
\end{subfigure}\hfill
\begin{subfigure}[t]{0.30\linewidth}
  \centering
  \includegraphics[width=\linewidth]{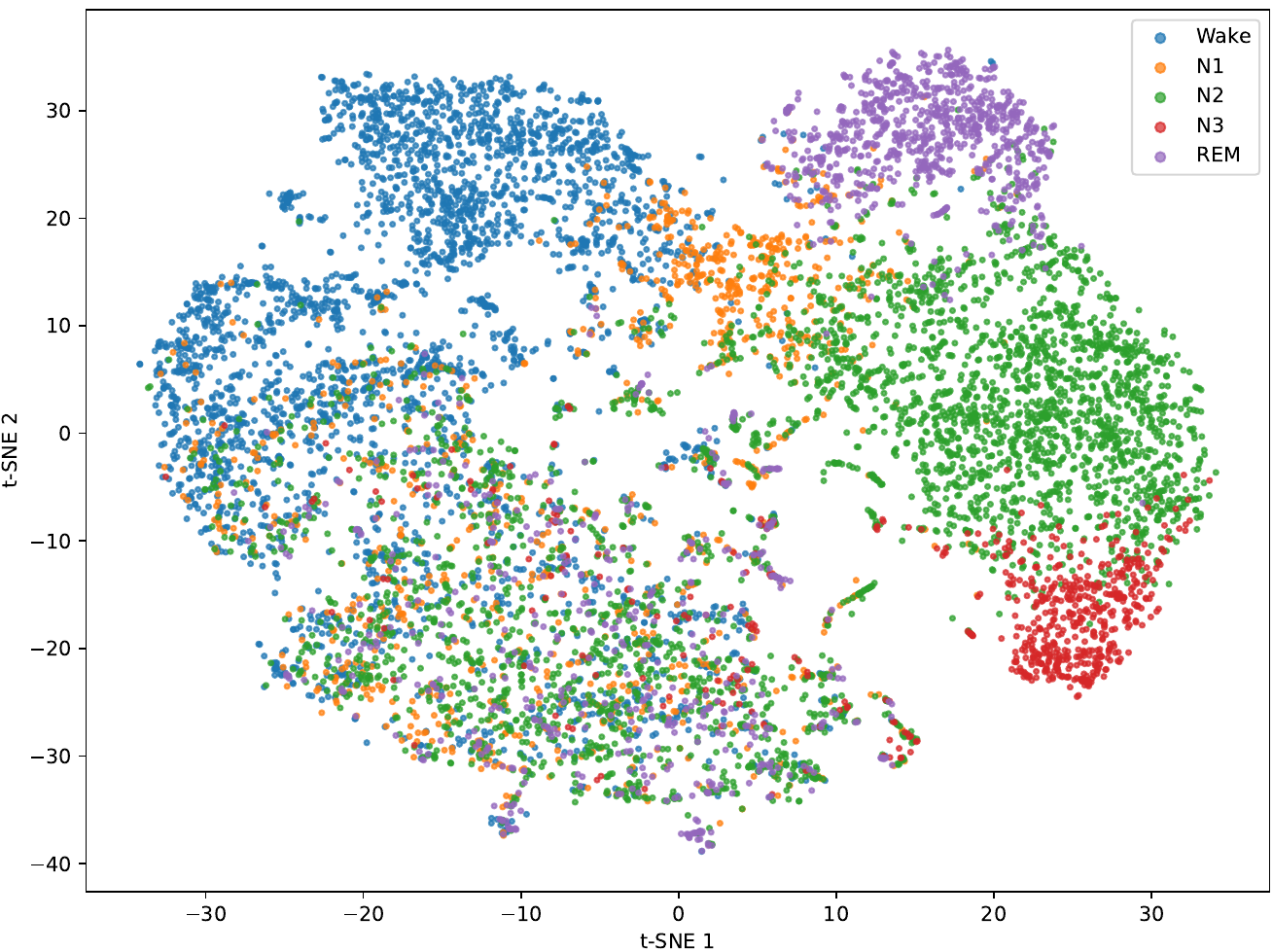}
  \caption{SleepEDF-20}
\end{subfigure}

\par\vspace*{5pt}

\begin{subfigure}[t]{0.30\linewidth}
  \centering
  \includegraphics[width=\linewidth]{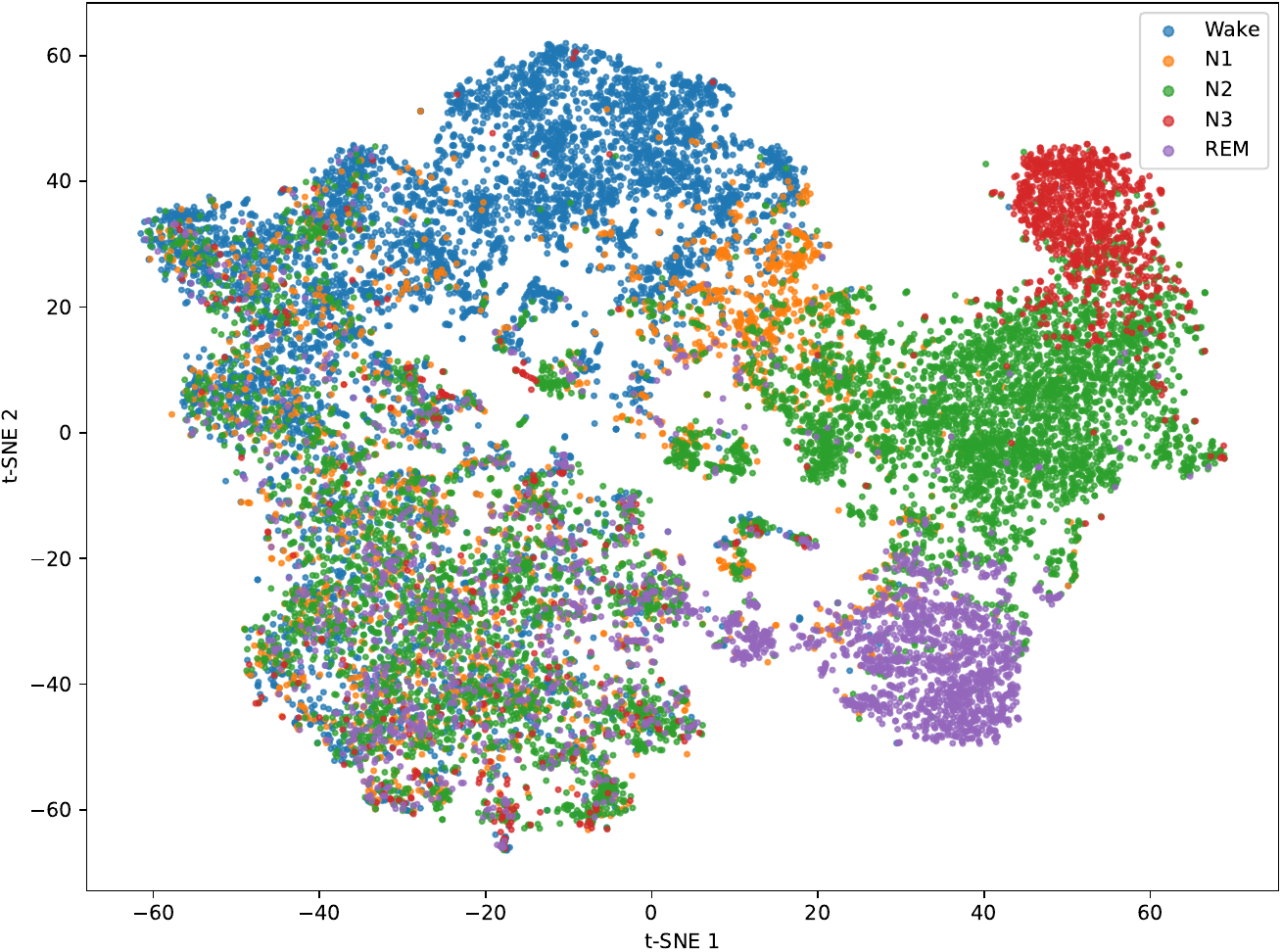}
  \caption{SHHS}
\end{subfigure}\hfill
\begin{subfigure}[t]{0.30\linewidth}
  \centering
  \includegraphics[width=\linewidth]{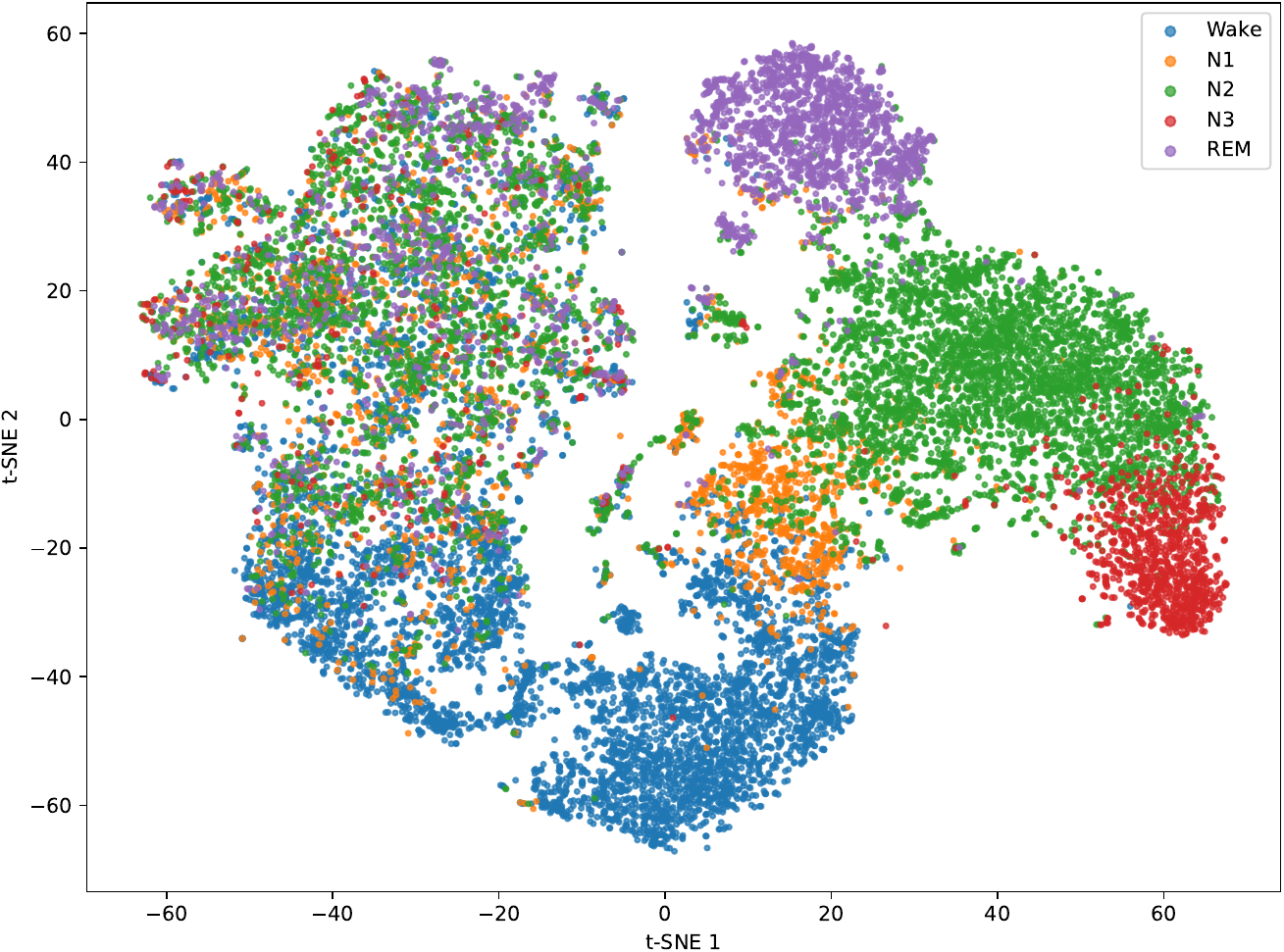}
  \caption{EDF-ISRUC-S3}
\end{subfigure}\hfill
\begin{subfigure}[t]{0.30\linewidth}
  \centering
  \includegraphics[width=\linewidth]{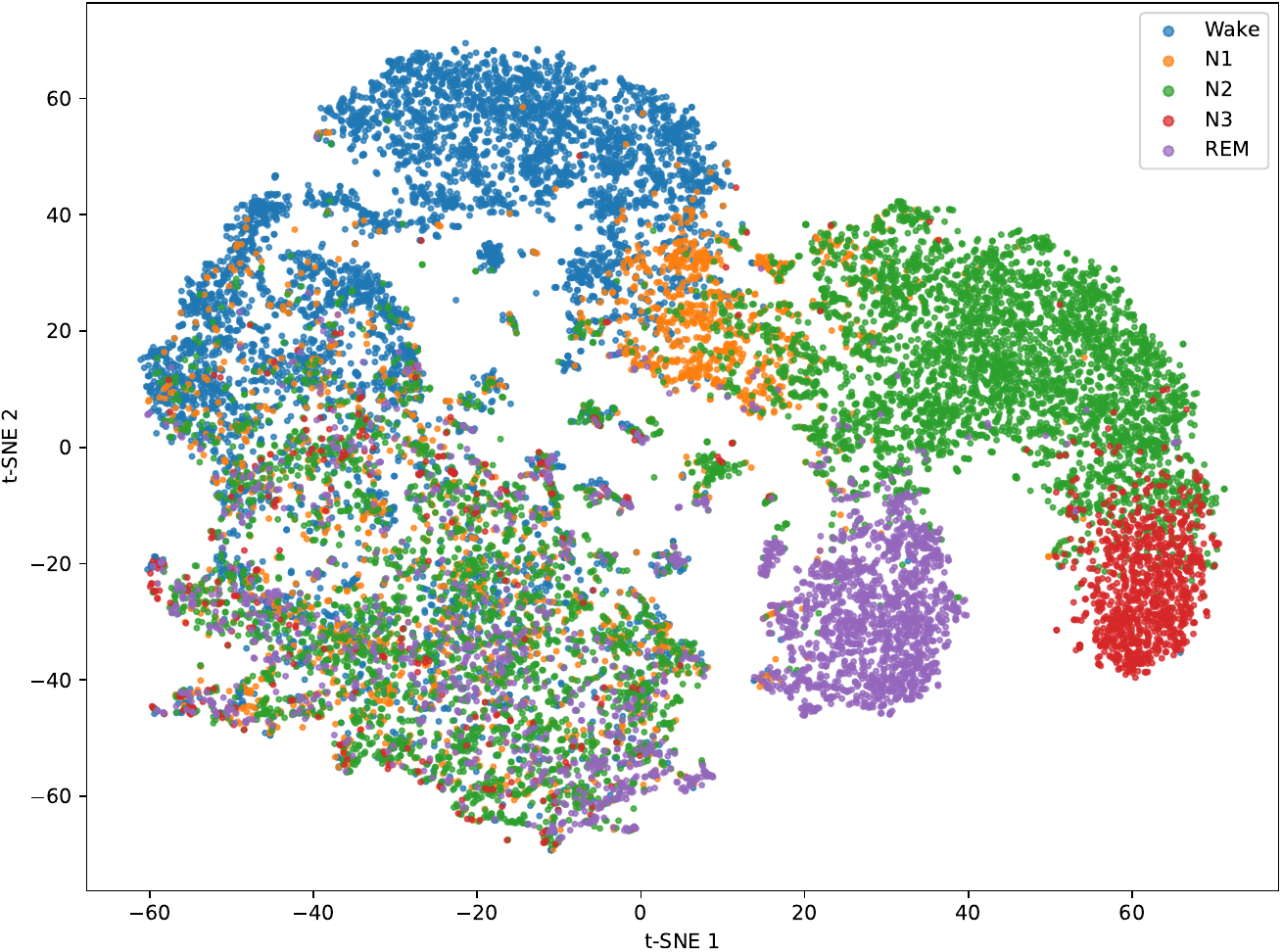}
  \caption{SleepEDF-78}
\end{subfigure}

\caption{t-SNE visualizations of feature representations learned by a single multi-channel model. Each point corresponds to one EEG epoch and is colored according to the sleep stage. Subfigures (A–D) show results on four individual datasets, while (E–F) present cross-dataset settings. The learned representations exhibit compact intra-class clustering and consistent inter-class separation across datasets, indicating stable and generalizable feature learning.}
\label{fig.6}
\end{figure*}

\renewcommand{\thesubsection}{\Alph{subsection}}
\setcounter{subsection}{0}   

\subsection{Parameter Sensitivity}
As shown in Fig.~\ref{fig.4}, based on the aforementioned experimental results, we further conducted a systematic analysis of the stability and robustness of the proposed framework under different parameter configurations, with a particular focus on examining the impact of several key hyperparameters on model performance.

\subsection{Confusion Matrices}
\label{sec:ablation}

As shown in Fig.~\ref{fig.5}, the confusion matrices across SleepEDF-78, SleepEDF-20, SHHS, and ISRUC-S3 consistently exhibit strong diagonal dominance, indicating that the proposed model achieves reliable and stable sleep stage discrimination under diverse recording conditions. Wake, N3, and REM stages are recognized with high accuracy, whereas misclassifications mainly occur between physiologically ambiguous neighboring stages, particularly N1 and N2, which remains a well-known challenge in automatic sleep staging.  Importantly, under strict cross-dataset evaluations (SleepEDF-78 $\rightarrow$ SHHS / ISRUC-S3), the overall prediction patterns remain highly comparable to those observed in single-dataset settings, without evident systematic bias and with only limited performance degradation. These results suggest that the proposed model learns dataset-invariant discriminative representations and demonstrates strong robustness and generalization capability across heterogeneous cohorts.

\captionsetup{belowskip=2pt}
\subsection{t-SNE Plots}
As shown in Fig.~\ref{fig.6}, the t-SNE visualizations illustrate that samples belonging to the same sleep stage form compact and well-separated clusters in the learned feature space, indicating that the proposed model captures discriminative representations aligned with accurate sleep stage classification. This clustering behavior is consistently observed across all four datasets (SleepEDF-78, SleepEDF-20, SHHS, and ISRUC-S3), demonstrating stable and coherent representation learning under diverse data distributions.
Importantly, under cross-dataset evaluation settings (SleepEDF-78 $\rightarrow$ SHHS / ISRUC-S3), the overall cluster structures are largely preserved, without evident cluster collapse or pronounced inter-class mixing. This observation suggests that the learned feature representations remain stable and discriminative under domain shifts, highlighting the robustness and cross-dataset generalization capability of the proposed model.

\subsection{Ablation Studies}
\subsubsection{Effect of Data-domain Filtering}

\begin{table}[htbp]
\centering
\caption{Ablation Study of Data-Domain Filtering}
\label{tab.5}
\small
\setlength{\tabcolsep}{5pt}
\renewcommand{\arraystretch}{1.2}

\begin{tabular}{@{}l ccc ccc@{}}
\toprule
\multirow{2}{*}{\textbf{Data Prepare}} 
& \multicolumn{3}{c}{\textbf{SleepEDF-78}} 
& \multicolumn{3}{c}{\textbf{SleepEDF-20}} \\
\cmidrule(lr){2-4} \cmidrule(lr){5-7}
 & \textbf{ACC} & \textbf{MF1} & $\boldsymbol{\kappa}$ 
 & \textbf{ACC} & \textbf{MF1} & $\boldsymbol{\kappa}$ \\
\midrule
Wavelet + STFT     & 86.6 & 83.2 & 0.827 & 88.5 & 84.8 & 0.841 \\
STFT               & 86.4 & 83.1 & 0.818 & 88.3 & 84.4 & 0.841 \\
\textbf{Diffusion + STFT}
                   & \textbf{88.0} & \textbf{84.2} & \textbf{0.836}
                   & \textbf{89.2} & \textbf{85.5} & \textbf{0.852} \\
\bottomrule
\end{tabular}
\end{table}

\begin{table}[htbp]
\centering
\caption{Denoising Performance Comparison on EEG Datasets}
\label{tab.6}
\small
\setlength{\tabcolsep}{2.5pt}
\renewcommand{\arraystretch}{0.95}

\begin{tabular}{@{}l l c c c c c@{}}
\toprule
\multicolumn{7}{c}{\textbf{EEGDenoise}} \\
\midrule
\textbf{Method} &  & \textbf{RRMSE} $\downarrow$ & \textbf{CC} $\uparrow$ & \textbf{SNR} $\uparrow$ & \textbf{WSNR} $\uparrow$ & \textbf{WCC} $\uparrow$ \\
\midrule
1D-ResCNN [39]     &  & 0.302 & 0.931 & 12.746 & 16.897 & 0.906 \\
EEGdenoiseNet [40]  &  & 0.344 & 0.904 & 12.121 & 15.094 & 0.888 \\
EEGDnet [41]      &  & 0.301 & 0.945 & 13.251 & 17.584 & 0.924 \\
DuoCL [42]         &  & 0.315 & 0.919 & 13.236 & 16.374 & 0.902 \\
GCTNet [43]        &  & 0.262 & 0.948 & 14.534 & 18.702 & 0.926 \\
\textbf{Ours}  &  & \textbf{0.257} & \textbf{0.945} & \textbf{14.513} & \textbf{18.787} & \textbf{0.933} \\
\midrule
\multicolumn{7}{c}{\textbf{MIT-BIH}} \\
\midrule
\textbf{Method} &  & \textbf{RRMSE} $\downarrow$ & \textbf{CC} $\uparrow$ & \textbf{SNR} $\uparrow$ & \textbf{WSNR} $\uparrow$ & \textbf{WCC} $\uparrow$ \\
\midrule
1D-ResCNN [39]     &  & 0.326 & 0.935 & 11.114 & 14.075 & 0.936 \\
EEGdenoiseNet [40]  &  & 0.302 & 0.933 & 12.912 & 15.698 & 0.938 \\
EEGDnet [41]       &  & 0.323 & 0.939 & 11.204 & 13.767 & 0.941 \\
DuoCL [42]         &  & 0.283 & 0.946 & 13.428 & 16.259 & 0.950 \\
GCTNet [43]        &  & 0.254 & 0.958 & 13.975 & 17.422 & 0.958 \\
\textbf{Ours}  &  & \textbf{0.251} & \textbf{0.968} & \textbf{14.087} & \textbf{17.523} & \textbf{0.964} \\
\bottomrule
\end{tabular}
\end{table}

To evaluate the effectiveness of the proposed data-domain stabilization strategy, we compare three data preprocessing schemes under identical network architectures and training configurations, including STFT, Wavelet+STFT, and Diffusion+STFT, with the results summarized in Table~\ref{tab.5}. On the SleepEDF-78 dataset, Diffusion+STFT achieves the best performance across all evaluation metrics, improving ACC, MF1, and Cohen’s $\kappa$ by 1.6\%, 1.1\%, and 0.018 over STFT alone, and yielding further gains of 1.4\%, 1.0\%, and 0.009 compared with Wavelet+STFT, respectively. Similar trends are observed on the SleepEDF-20 dataset, where Diffusion+STFT consistently outperforms the STFT-only baseline with improvements of 0.9\%, 1.1\%, and 0.011 in ACC, MF1, and $\kappa$. These results demonstrate that diffusion-based data-domain stabilization is more effective than conventional filtering or no-filtering strategies in enhancing signal quality, and provides stable and consistent performance improvements across datasets of different scales.

To further demonstrate the effectiveness of the proposed data-domain stabilization module in signal enhancement tasks, Table~\ref{tab.6} reports quantitative comparisons on two public benchmark datasets, EEGDenoise and MIT-BIH, using widely adopted signal quality metrics, including relative root mean square error (RRMSE), correlation coefficient (CC), signal-to-noise ratio (SNR), weighted signal-to-noise ratio (WSNR), and weighted correlation coefficient (WCC). As shown in Table~\ref{tab.6}, the proposed method consistently outperforms competing approaches across most metrics on both datasets. In particular, lower RRMSE values indicate more effective noise suppression in terms of signal amplitude reconstruction, while improvements in correlation-based metrics (CC and WCC) suggest better preservation of waveform morphology and phase structure. Moreover, gains in energy-related metrics (SNR and WSNR) demonstrate that the proposed approach effectively reduces noise energy across both global and weighted frequency bands. These results provide complementary evidence supporting the main experimental findings, confirming that the data-domain stabilization module can robustly suppress noise while preserving essential physiological signal structures under diverse noise conditions and across multiple datasets.

\begin{table}[htbp]
\centering
\caption{Overall Ablation of the Two-Stage Strategy Diffusion}
\label{tab.7}
\small
\setlength{\tabcolsep}{3pt}
\renewcommand{\arraystretch}{0.95}
\begin{tabular}{l l l ccc ccc}
\toprule
\multirow{2}{*}{\textbf{Baseline}} &
\multirow{2}{*}{\textbf{Feature}} &
\multirow{2}{*}{\textbf{Data}} &
\multicolumn{3}{c}{\textbf{SleepEDF-78}} &
\multicolumn{3}{c}{\textbf{SleepEDF-20}} \\
\cmidrule(lr){4-6} \cmidrule(lr){7-9}
 & & & \textbf{ACC} & \textbf{MF1} & $\boldsymbol{\kappa}$ & \textbf{ACC} & \textbf{MF1} & $\boldsymbol{\kappa}$ \\
\midrule
\checkmark & \checkmark & \checkmark & 88.0 & 84.2 & 0.836 & 89.2 & 85.5 & 0.852 \\
\checkmark & \checkmark & $\times$    & 87.4 & 83.5 & 0.824 & 88.5 & 85.1 & 0.849 \\
\checkmark & $\times$    & \checkmark & 86.3 & 82.4 & 0.806 & 87.1 & 84.3 & 0.836 \\
\checkmark & $\times$    & $\times$    & 84.6 & 81.1 & 0.789 & 85.9 & 82.8 & 0.819 \\
\bottomrule
\end{tabular}
\end{table}
\subsubsection{Effect of the Two-stage Strategy}
To verify the overall effectiveness of the proposed two-stage diffusion strategy, we conduct an ablation study under a unified baseline model by selectively enabling or disabling the data-domain diffusion and feature-domain diffusion modules. The experimental results are summarized in Table~\ref{tab.7}.

On the SleepEDF-78 dataset, introducing either feature-domain diffusion or data-domain diffusion alone leads to performance improvements over the baseline model, although the magnitude of the gains differs noticeably. When both diffusion strategies are jointly applied, the model achieves the best overall performance, with ACC, MF1, and $\kappa$ reaching 88.0\%, 84.2\%, and 0.836, respectively. Compared with using only feature-domain diffusion, the full model further improves ACC, MF1, and $\kappa$ by 0.6\%, 0.7\%, and 0.012, respectively. The improvements are even more pronounced when compared with using only data-domain diffusion, with gains of 1.7\%, 1.8\%, and 0.030. When both diffusion strategies are removed, the model performance degrades substantially, with ACC and MF1 decreasing by 3.4\% and 3.1\%, respectively.

A consistent trend is observed on the SleepEDF-20 dataset. Both data-domain diffusion and feature-domain diffusion individually enhance the baseline performance, while their combination consistently yields the best results. Relative to using only feature-domain diffusion, the complete two-stage strategy improves ACC, MF1, and $\kappa$ by 0.7\%, 0.4\%, and 0.003, respectively. Compared with using only data-domain diffusion, the corresponding improvements are 2.1\%, 1.2\%, and 0.016. When only the baseline model is retained, the overall performance exhibits the most significant degradation.

These results indicate that data-domain diffusion and feature-domain diffusion do not contribute in a merely additive manner. The former primarily enhances the discriminability of the input signals, while the latter further refines the feature representations. By operating at different levels, the two diffusion stages complement each other. Through this coordinated two-stage design, DSSNet achieves consistent and robust improvements in overall sleep staging performance.

\subsubsection{Effect of Feature Layers}

\begin{table}[htbp]
\centering
\caption{Layer Ablation Study in the Feature Domain}
\label{tab.8}
\small
\renewcommand{\arraystretch}{1.2}
\setlength{\tabcolsep}{3.5pt}
\begin{tabular}{l ccc ccc}
\toprule
\multirow{2}{*}{\textbf{Layer Ablation}} 
& \multicolumn{3}{c}{\textbf{SleepEDF-78}} 
& \multicolumn{3}{c}{\textbf{SleepEDF-20}} \\
\cmidrule(lr){2-4} \cmidrule(lr){5-7}
 & \textbf{ACC} & \textbf{MF1} & $\boldsymbol{\kappa}$ 
 & \textbf{ACC} & \textbf{MF1} & $\boldsymbol{\kappa}$ \\
\midrule
L1 + L2 + L3 + L4 & 88.0 & 84.2 & 0.836 & 89.2 & 85.5 & 0.852 \\
L2 + L4           & 87.4 & 83.6 & 0.814 & 88.6 & 84.9 & 0.848 \\
L4                & 86.9 & 83.1 & 0.807 & 88.2 & 84.5 & 0.843 \\
\bottomrule
\end{tabular}
\end{table}

To analyze the effect of feature-domain diffusion stabilization at different network hierarchies, we conduct an ablation study on the SleepEDF-78 and SleepEDF-20 datasets by varying the deployment of the diffusion modules across feature levels. Specifically, we consider three settings: applying feature-domain diffusion at all feature layers (L1+L2+L3+L4), at a subset of layers (L2+L4), and only at the deepest layer (L4). The corresponding results are reported in Table~\ref{tab.8}.

On the SleepEDF-78 dataset, when feature-domain diffusion is applied to all feature layers (L1--L4), the model achieves an ACC of 88.0\%, an MF1 of 84.2\%, and a $\kappa$ value of 0.836. When diffusion is deployed only at partial layers (L2+L4), the performance drops to 87.4\% / 83.6\% / 0.814. Further restricting diffusion stabilization to only the deepest layer (L4) leads to a more pronounced decline, with performance decreasing to 86.9\% / 83.1\% / 0.807. A consistent downward trend across all three metrics is observed as the coverage of diffusion across feature layers is reduced.

A similar pattern is observed on the SleepEDF-20 dataset. With diffusion applied at all layers (L1--L4), the model achieves 89.2\% ACC, 85.5\% MF1, and a $\kappa$ of 0.852. When diffusion is introduced only at L2+L4, the performance decreases to 88.6\% / 84.9\% / 0.848. When diffusion is applied exclusively at L4, the performance further declines to 88.2\% / 84.5\% / 0.843.

These results indicate that the performance gains from feature-domain diffusion stabilization depend on the extent of its deployment across network layers. Compared with introducing stabilization constraints at a single layer or a limited number of higher-level features, enforcing diffusion-based stabilization across multiple feature hierarchies more effectively suppresses representation drift at different abstraction levels, resulting in more stable and consistent performance improvements on both datasets.

\subsection{Multi-channel Analysis}

\begin{table}[htbp]
\centering
\caption{Overall Ablation of Data Channels}
\label{tab.9}
\small
\setlength{\tabcolsep}{4pt}
\renewcommand{\arraystretch}{0.95}
\begin{tabular}{l l l ccc ccc}
\toprule
\multirow{2}{*}{\textbf{EEG}} &
\multirow{2}{*}{\textbf{EOG}} &
\multirow{2}{*}{\textbf{EMG}} &
\multicolumn{3}{c}{\textbf{SleepEDF-78}} &
\multicolumn{3}{c}{\textbf{SleepEDF-20}} \\
\cmidrule(lr){4-6} \cmidrule(lr){7-9}
 & & & \textbf{ACC} & \textbf{MF1} & $\boldsymbol{\kappa}$ 
       & \textbf{ACC} & \textbf{MF1} & $\boldsymbol{\kappa}$ \\
\midrule
\checkmark & \checkmark & \checkmark & 88.0 & 84.2 & 0.836 & 89.2 & 85.5 & 0.852 \\
\checkmark & \checkmark & $\times$    & 87.1 & 82.8 & 0.803 & 88.8 & 84.2 & 0.837 \\
\checkmark & $\times$    & $\times$    & 85.4 & 81.4 & 0.782 & 87.7 & 82.6 & 0.818 \\
\bottomrule
\end{tabular}
\end{table}

To investigate the contribution of different physiological signal channels, we conduct ablation experiments on the SleepEDF-78 and SleepEDF-20 datasets by varying the input configurations, including EEG+EOG+EMG, EEG+EOG, and EEG only. The quantitative results are summarized in Table~\ref{tab.9}.

On the SleepEDF-78 dataset, the full three-channel configuration achieves an ACC / MF1 / $\kappa$ of 88.0\% / 84.2\% / 0.836. Removing the EMG channel leads to a performance decrease to 87.1\% / 82.8\% / 0.803, and further degradation is observed when only EEG is used, yielding 85.4\% / 81.4\% / 0.782. A similar trend is observed on SleepEDF-20, where the three-channel configuration reaches 89.2\% / 85.5\% / 0.852, the EEG+EOG setting achieves 88.8\% / 84.2\% / 0.837, and the EEG-only configuration results in 87.7\% / 82.6\% / 0.818.

These results indicate that EEG provides the primary discriminative information for sleep stage classification, while the inclusion of EOG and EMG signals further enhances overall performance and classification consistency. Moreover, multi-channel inputs consistently demonstrate superior robustness across datasets, highlighting the benefit of leveraging complementary physiological information for stable sleep staging.

\subsection{Computational Efficiency Analysis}

\begin{table}[!t]
\centering
\caption{Computational Cost under Different Feature-Layer Configurations}
\label{tab.10}

\footnotesize
\setlength{\tabcolsep}{16pt}
\renewcommand{\arraystretch}{1.25}

\begin{tabular}{@{}c c c c@{}}
\toprule

\multicolumn{4}{c}{\textbf{L1+L2+L3+L4}} \\
\midrule
\textbf{Stage} & \textbf{Params (M)} & \textbf{FLOPs (G)} & \textbf{FPS (samples/s)} \\
\midrule
1 & 9.276 & 1.883 & 1265.03 \\
2 & 9.276 & 2.508 & 1015.05 \\
3 & 9.276 & 3.133 & 726.62  \\
4 & 9.276 & 3.758 & 672.28  \\
\midrule

\multicolumn{4}{c}{\textbf{L2+L4}} \\
\midrule
\textbf{Stage} & \textbf{Params (M)} & \textbf{FLOPs (G)} & \textbf{FPS (samples/s)} \\
\midrule
1 & 7.828 & 1.569 & 1436.27 \\
2 & 7.828 & 1.882 & 1145.35 \\
3 & 7.828 & 2.194 & 1129.07 \\
4 & 7.828 & 2.507 & 845.18  \\
\midrule

\multicolumn{4}{c}{\textbf{L4}} \\
\midrule
\textbf{Stage} & \textbf{Params (M)} & \textbf{FLOPs (G)} & \textbf{FPS (samples/s)} \\
\midrule
1 & 7.103 & 1.412 & 1420.07 \\
2 & 7.103 & 1.569 & 1439.92 \\
3 & 7.103 & 1.725 & 1353.32 \\
4 & 7.103 & 1.881 & 1281.83 \\

\bottomrule
\end{tabular}
\end{table}

To analyze the computational overhead introduced by different feature-domain diffusion configurations, we evaluate the model parameter size, computational complexity (FLOPs), and inference efficiency (FPS) under three settings: L1+L2+L3+L4, L2+L4, and L4. The quantitative results are summarized in Table~\ref{tab.10}.

As shown in Table~\ref{tab.10}, the total number of model parameters remains largely consistent across different configurations, whereas the computational overhead is primarily reflected in variations in FLOPs and inference speed. Under full-layer diffusion (L1+L2+L3+L4), the FLOPs increase progressively with network depth (from 1.883G to 3.758G), accompanied by a notable reduction in inference speed (from 1265.03 to 672.28 samples/s). In contrast, introducing diffusion at partial layers (L2+L4) or exclusively at the deepest layer (L4) substantially reduces computational cost and correspondingly improves inference efficiency. Among the evaluated configurations, the L4 setting achieves the lowest FLOPs and the highest inference speed across all stages.

Overall, these results indicate that feature-domain diffusion primarily increases computational complexity rather than model capacity, as the parameter size remains largely unchanged. Moreover, the computational overhead grows with the number of diffusion-enhanced layers. This observation suggests that a practical trade-off can be achieved between performance gains and computational efficiency by selectively applying diffusion to deeper feature layers, depending on deployment constraints and available computational resources.

\section{Discussion}

The experimental results demonstrate that data-domain and feature-domain diffusion play distinct yet complementary roles within the proposed two-stage stabilization framework. Data-domain diffusion primarily aims to recover stable and physiologically meaningful signal structures and is able to deliver clear performance benefits with a relatively small number of inference steps. In contrast, feature-domain diffusion mainly addresses representation drift across subjects and datasets, thereby exerting a more direct influence on the robustness and generalization capability of the learned features.

At the same time, the diffusion-based inference process introduces an inherent trade-off between performance improvement and computational efficiency. In the data domain, excessively long diffusion trajectories lead to a substantial increase in computational cost with diminishing performance returns. In the feature domain, overly aggressive multi-step corrections may inadvertently weaken the discriminative power of high-level semantic representations. These observations indicate that the current framework relies on empirically determined diffusion configurations to balance stability and discriminability.

Building on these findings, future work may explore adaptive diffusion inference strategies that are guided by signal quality or feature uncertainty, enabling dynamic adjustment of inference step lengths and correction strengths across different network stages. Such strategies could achieve a more favorable balance between performance and efficiency. Furthermore, integrating feature-domain diffusion with cross-layer modeling or multi-scale consistency constraints represents a promising direction for further enhancing robustness under complex cross-domain distribution shifts.

\end{document}